\documentclass[letterpaper, 10 pt, conference]{ieeeconf}  

\IEEEoverridecommandlockouts                              

\usepackage{amsmath} 
\usepackage{amssymb}  
\usepackage{gensymb}
\usepackage{graphicx}
\usepackage{epstopdf}
\usepackage{subcaption}
\usepackage[colorlinks]{hyperref}
\usepackage{multirow}
\usepackage[T1]{fontenc}
\usepackage{fix-cm}

\makeatletter
\let\NAT@parse\undefined
\makeatother

\usepackage[square,comma,numbers,sort&compress]{natbib} 

\title{\LARGE \bf
SPLODE: Semi-Probabilistic Point and Line Odometry with Depth Estimation from RGB-D Camera Motion
}

\author{Pedro F. Proen\c{c}a$^{1}$ and Yang Gao$^{1}$
\thanks{$^{1}$The authors are with the Surrey Space Centre, Faculty of Engineering and Physical Sciences, University of Surrey, GU2 7XH Guildford,
	U.K. {\tt\small \{p.proenca, yang.gao\}@surrey.ac.uk}}%
}

\begin{document}

\maketitle
\thispagestyle{empty}
\pagestyle{empty}

\begin{abstract}
Active depth cameras suffer from several limitations, which cause incomplete and noisy depth maps, and may consequently affect the performance of RGB-D Odometry.
To address this issue, this paper presents a visual odometry method based on point and line features that leverages both measurements from a depth sensor and depth estimates from camera motion. Depth estimates are generated continuously by a probabilistic depth estimation framework for both types of features to compensate for the lack of depth measurements and inaccurate feature depth associations. The framework models explicitly the uncertainty of triangulating depth from both point and line observations to validate and obtain precise estimates.\par
Furthermore, depth measurements are exploited by propagating them through a depth map registration module and using a frame-to-frame motion estimation method that considers 3D-to-2D and 2D-to-3D reprojection errors, independently. Results on RGB-D sequences captured on large indoor and outdoor scenes, where depth sensor limitations are critical, show that the combination of depth measurements and estimates through our approach is able to overcome the absence and inaccuracy of depth measurements.
\end{abstract}


\section{Introduction}

Modern active depth cameras, i.e., structured-light and time-of-flight (ToF) cameras, are capable of capturing dense 640$\times$480 depth maps from poorly textured scenes, at 30 fps. Therefore, the combination of these depth cameras with colour cameras, in a stereo configuration, has become a popular setup to capture both the geometry and appearance of indoor scenes. The calibration of such stereo setup allows the direct registration of depth with RGB images. To leverage this type of fused data, known as RGB-D data, many visual odometry and SLAM methods \cite{RGBD_Switch,DEMO,pinpointSLAM,henry2012,DVO_2,PLVO,whelan2013fusion} have emerged, over the last few years.
However, these depth sensors suffer from several limitations, which cause missing and erroneous depth readings, such as: limited field of view and range (typically between 0.4 and 5 m), near-infrared (NIR) interference (e.g. sunlight and multiple devices) and non-Lambertian reflections. These limitations may affect severely the performance of pure RGB-D based methods \cite{henry2012,DVO_2,PLVO,whelan2013fusion} as these only make use of RGB pixels that have associated depth measurements. On the other hand, depth estimation from a single camera motion is less limited by the above issues but purely monocular SLAM \cite{engel2013semi, SVO} suffers from scale drift and degenerate motion. \par

\begin{figure}[t]
\centering
	\begin{tabular}{@{}c@{ }c@{}}
		\includegraphics[scale=0.25]{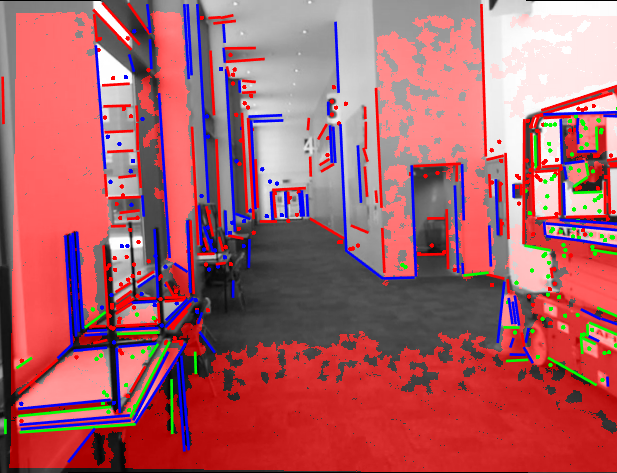} &
		\includegraphics[scale=0.25]{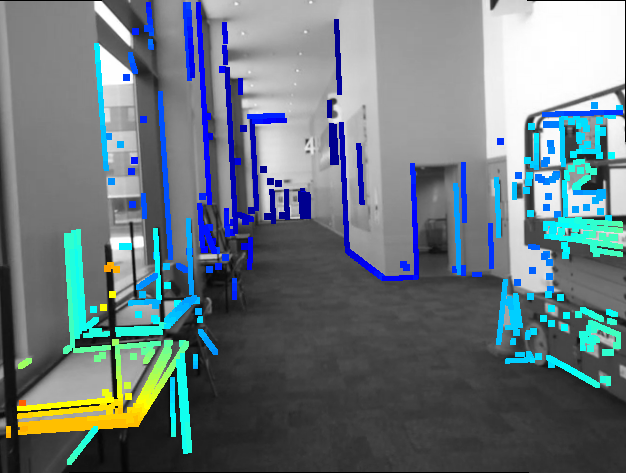}\\
		\includegraphics[scale=0.25]{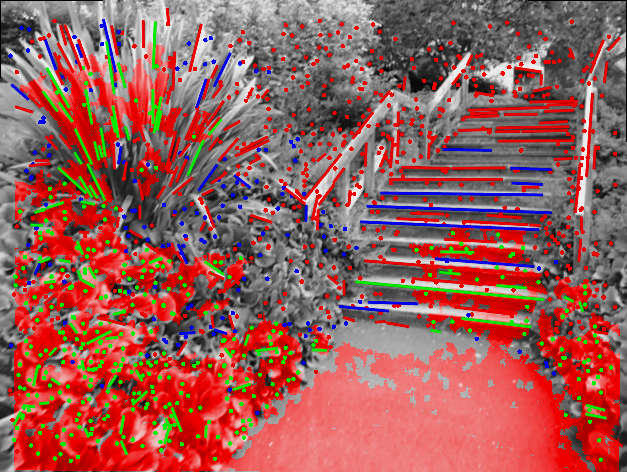} &
		\includegraphics[scale=0.25]{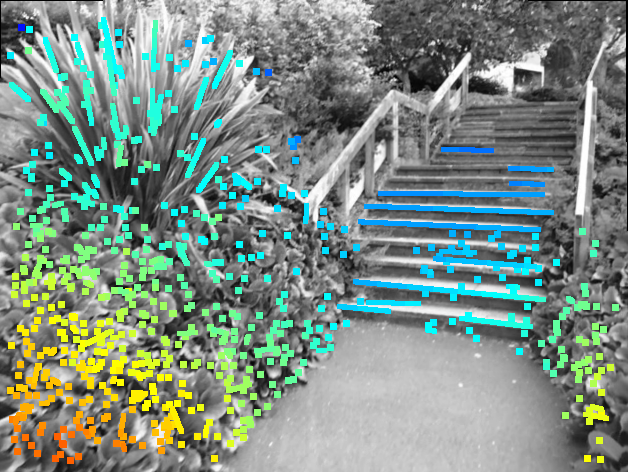}
	\end{tabular} 
	\caption{Critical indoor and outdoor scenes for active depth cameras. The depth maps captured by a structured-light sensor are mapped and coloured in red, on the left images. Our method exploits points and lines with associated depth measurements from a depth camera and depth estimates from temporal stereo, shown respectively on the left image in green and blue. The respective inverse depth is shown (colour-coded) on the right for both measurements and estimates. Red features do not contain valid depth.}
	\vspace*{-1mm} 
	\label{fig1}
\end{figure}

\begin{figure*}
\centering
		\includegraphics[scale=0.45]{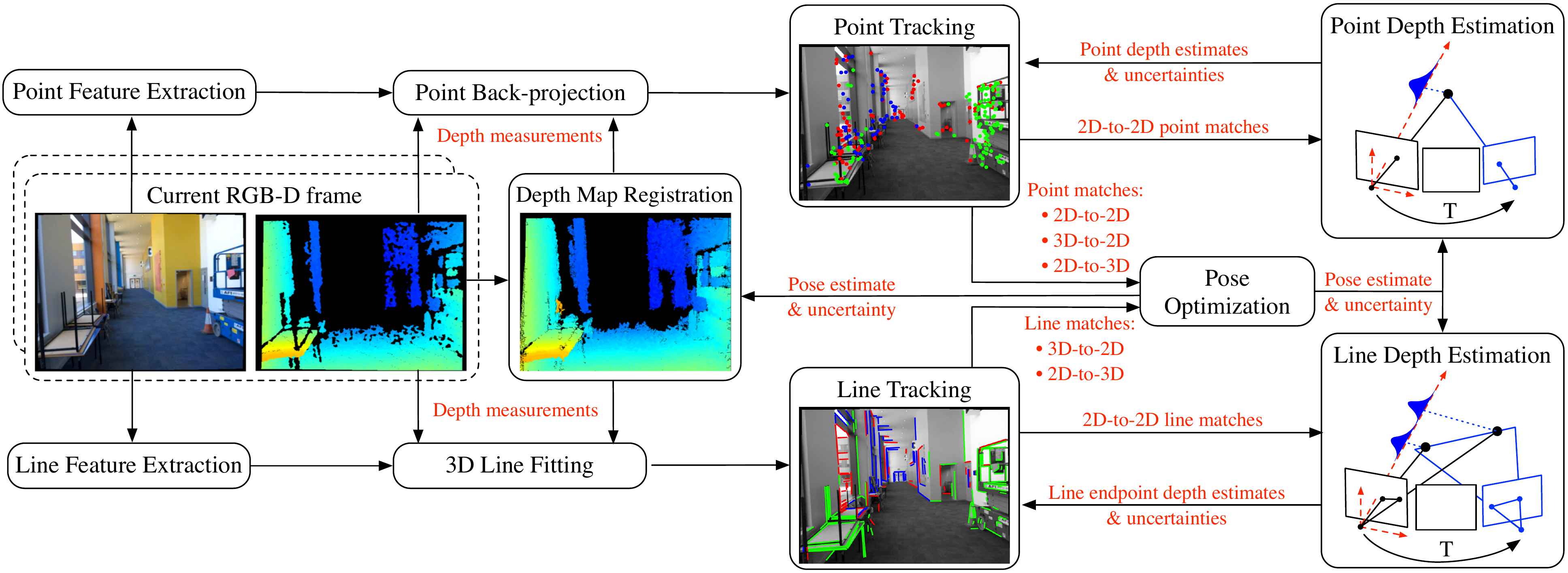}
	\caption{Overview of SPLODE system.}
	\label{fig2}
	\vspace*{-1mm} 
\end{figure*}

To address these limitations, we propose a robust visual odometry based on image point and line features, called SPLODE, which exploits not only the depth measurements from a depth camera but also depth estimates from camera motion. While SPLODE overcomes temporally unavailable depth measurements by propagating depth through depth map registration and estimating frame-to-frame pose by using loosely both the 2D-to-3D and 3D-to-2D reprojection errors of feature matches, SPLODE continuously estimates depth for tracked features by recursively triangulating and fusing depth from temporal stereo. The key contributions of this paper are the following: 
\begin{itemize}
\item The introduction of a depth estimation framework that models explicitly the triangulation uncertainty of points and lines by taking into account their particular geometries and the pose uncertainty.
\item Showing how to minimize the impact of depth measurement errors on the pose estimation by comparing two different depth estimation schemes.
\end{itemize}
Our experiments show that the combination of depth camera measurements with estimates through the proposed framework allows SPLODE to operate on challenging environments (i.e. beyond room-sized environments), where the limitations of depth cameras become too problematic to rely just on RGB-D odometry.

\section{Related Work}

Recently, several works have shown the benefit of combining different depth sensing modalities for ego-motion estimation. Stereo cameras have been integrated in monocular SLAM methods based on either direct image alignment \cite{stereo_lsdslam} or point features \cite{ORBSLAM2} to help resolving scale-ambiguity and degenerate motions. RGB-D cameras have been used with monocular techniques in \cite{RGBD_Switch, DEMO, pinpointSLAM} to deal with a lack of depth sensor measurements. Hu et al. \cite{RGBD_Switch} proposed a heuristic switching algorithm to choose between RGB-D SLAM and monocular SLAM based on epipolar geometry, however with such a hard switch, all depth measurements during the monocular SLAM mode are ignored, thus a map optimization is required to estimate the scale. Zhang et al. \cite{DEMO} proposed estimating motion by combining 3D-to-2D and 2D-to-2D feature point matches in a joint optimization. Additionally, a depth map registration module is proposed to incorporate depth information from past frames and extend the method to LIDAR, nevertheless, during the frame-to-frame pose estimation, depth from the second frame (i.e. 2D-to-3D) is neglected. Alternatively, Ataer-Cansizoglu et al. \cite{pinpointSLAM} proposed using a RANSAC framework to combine hybrid features. The framework estimates pose by trying different triplets comprised of either plane matches and 3D-to-3D point matches or just 2D-to-3D point matches, whereas the rest of the matches, including point matches without depth, are used to check the hypothesis consensus, yet the case of 3D-to-2D matches is not considered. Both \cite{DEMO} and \cite{pinpointSLAM} triangulate points without depth after being tracked for longer than a certain temporal stereo baseline. However, assessing the stereo baseline, by itself, is not enough, especially for lines, e.g., camera translation parallel to a line or in a direction of a point renders triangulation impossible. \par
More convincing depth estimation is found among monocular odometry methods \cite{engel2013semi,SVO}. Engel et al. \cite{engel2013semi} estimated and refined a semi-dense inverse depth map for direct image alignment by modelling the inverse depth uncertainty of pixels with three criteria: An approximation to geometric disparity error, a photometric disparity error and a pixel to inverse depth conversion. Forster et al. \cite{SVO} introduced a fast monocular odometry, where the inverse depth of point features was modelled as a Gaussian $+$ Uniform mixture model by employing a Bayesian filter to fit good depth observations and account for outliers. Such model suggests robustness to repetitive textures, however, unlike our approach, the depth-filter needs to undergo multiple observations before it converges since it does not take into account the uncertainty of each feature observation. This method was further extended in \cite{PLSVO} to line features to address textureless scenes. \par
RGB-D data has also potential to be used in poorly textured and illuminated scenes due to the use of active light. In \cite{henry2012}, the combination of point features with ICP based on the depth data proved to be advantageous in dark areas. Similarly in \cite{DVO_2}, the direct minimization of both photometric and depth error showed better performance in low textured scenes than by just relying on photometric error. Robustness in RGB-D methods has also been achieved \cite{PointsAndPlanes_2013,PLVO} by exploiting other geometric primitives besides points. Planes have been widely used \cite{PointsAndPlanes_2013,pinpointSLAM}. More closely related to our work is the one of Lu and Song \cite{PLVO}, who combined points and line segment features under a framework that models the uncertainties of the 3D features based on the low-resolution and noise of depth measurements obtained by structured-light sensors. Despite their use of a 3D line fitting method that tolerates missing depth, the frame-to-frame motion estimation itself requires feature matches to have depth measurements in both frames, as it minimizes the Mahalanobis distance between the 3D feature locations. Consequently, the framework does not allow the inclusion of either features with missing depth or triangulated features.

\section{Method}
\label{sec:method}

\begin{figure}
\centering
\includegraphics[scale=0.45]{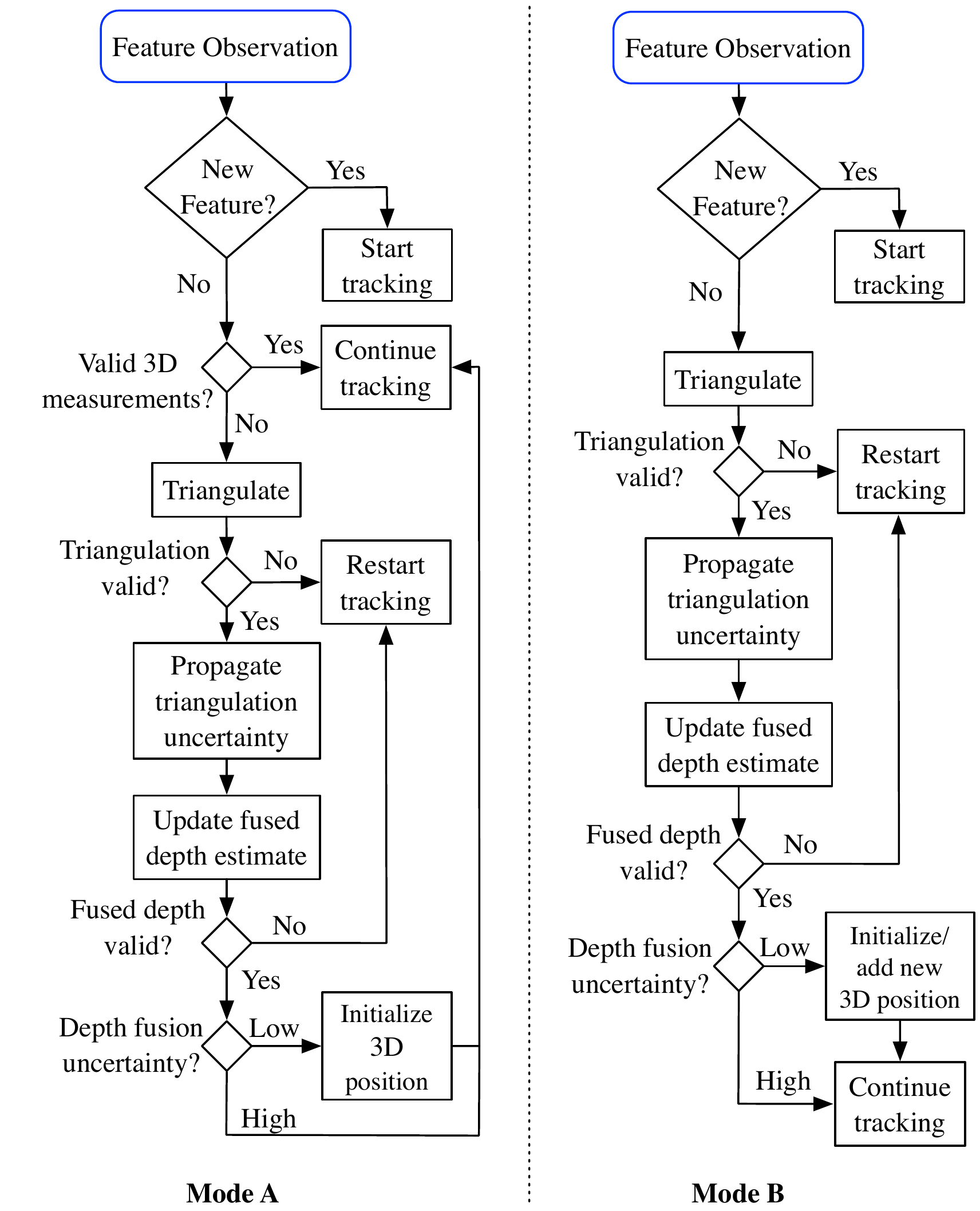}
	\caption{General feature workflow after matching and pose estimation for two proposed schemes to combine depth estimates and measurements: Mode A only estimates depth for features without valid depth measurements, whereas mode B estimates depth for all feature matches.}
	\vspace*{-3mm} 
	\label{fig3}
\end{figure}

Our proposed visual odometry system, outlined in Fig. \ref{fig2}, leverages active depth sensing and temporal stereo to retrieve the 3D location of points and line features, which are projected on the RGB camera. Once a new RGB-D frame is captured, 2D features are detected from the current RGB image and depth measurements are sampled from the current depth map to attempt to obtain the respective 3D coordinates. These features are then matched against the ones extracted from the last frame and pose is estimated by minimizing their bidirectional reprojection errors by employing an M-estimator to reduce the impact of spurious matches and wrong depth associations on the estimated pose. Given the pose estimate, past depth measurements can be combined with the current depth map via point cloud registration to achieve a denser depth map, from which depth is resampled and used to recover the 3D coordinates of features, which do not have yet valid depth,
so they can be used in the next frame-to-frame pose estimation. For the purpose of depth triangulation, features are tracked by frame-to-frame feature matching since their first frame observation to allow a wide-baseline stereo. As outlined by the feature tracking algorithm in Fig. \ref{fig3}, we have experimented two different schemes to integrate the depth estimation from temporal stereo:
\begin{itemize}
\item Mode A, similarly to \cite{DEMO, pinpointSLAM, ORBSLAM2}, only estimates depth for features that do not have associated depth measurements. As a result, the set of features with depth measurements and the set with depth estimates are disjoint.
\item Mode B estimates depth for features, regardless of having depth measurements, such that, even if a feature already has a 3D position, recovered from the depth map, a depth estimate may be applied to initialize a new 3D position hypothesis to be used along with the other hypothesis in the next frame's pose estimation as two 3D-to-2D feature match residuals. Since the pose estimation implements an M-estimator, the spurious depth hypothesis can be downweighted.\end{itemize}
While mode A gives priority to measurements coming from the depth sensor, estimating depth only when necessary, mode B relies less on depth measurements by treating the measurements and estimates equally. The principle behind mode B is that depth measurements may be inaccurate and depth estimates from temporal stereo can be used as an alternative hypothesis. The depth estimation module uses the uncertainty of the stereo triangulation, given the pose uncertainty, to fuse estimates from multiple views and assess the precision of the estimates. Whenever, the uncertainty of a fused depth estimate drops below a certain threshold, that depth estimate is applied to initialize the 3D position of the feature through back-projection, so it can be used in the next camera pose estimation. To combat wrong feature matches, the consistency of both triangulated and fused depth estimates are continuously checked based on the reprojection errors. In the remainder of this section, we detail each individual module of our framework.

\subsection{Preliminaries}
\label{sec:pre}
The projection of a 3D point: $P=\begin{bmatrix} X&Y&Z\end{bmatrix}^\top$ on an image plane is denoted as: $p = \pi(P)$ where the coordinates of this projected point: $p=\{x,y\}$ correspond to its homogeneous vector representation: $\dot{p}=P/Z$. The projected point $p$ can also be expressed in pixels by: $\begin{bmatrix} u&v&1\end{bmatrix}^\top=K\dot{p}$ where $K$ is known as the camera calibration matrix. \par

Both in 2D and 3D, a line segment is represented by its endpoints. For the purpose of measuring point-to-line distances, a 2D line is also parameterized in the Hessian normal form as: $l =\begin{bmatrix} l_x&l_y&d\end{bmatrix}$. \par
Furthermore, a point $P$ can be expressed with respect to a reference frame of a different camera view through a 3D rigid body transformation: $\{R,t \mid R \in SO(3), t \in \mathbb{R}^3\}$, such that: $P'=RP+t$. When necessary, rotation is expressed as a unit quaternion $q$ to achieve minimal pose parameters: $\xi=\{t_x,t_y,t_z,q_1,q_2,q_3\}$ such that $q_4 = \sqrt{1-q_1^2 - q_2^2 - q_3^2}$ for rotation angles between $[-\pi, \pi]$.

\subsection{Pose Estimation}
\label{sec:pose_est}
As several works in RGB-D Odometry \cite{DEMO, henry2012}, we avoid working in the Euclidean space, due to the depth sensor error at long distances, and instead estimate the frame-to-frame pose by jointly minimizing the reprojection errors of both line and point correspondences. To exploit the depth measurements available in both frames, the bidirectional error is used loosely, such that 3D-to-2D correspondences are used if depth is available on the first frame and 2D-to-3D correspondences are used if depth is available on the second frame. Given a 3D-to-2D point correspondence: $\{P,p'\}$, we express the reprojection error in the vector form as follows:
\begin{equation}
\label{eq:rep_err}
r_{p'}  = p' - \pi(\widehat{R}P +\hat{t}\, )
\end{equation}
where $\{\widehat{R},\hat{t}\,\}$ correspond to the relative pose estimate. Similarly, the reprojection error of a 2D-to-3D point match: $r_{p}$  can be expressed by using the inverse pose. For a 3D-to-2D correspondence of line segments, the residual is defined as the point-to-line distance between a 2D line: $l'$ and the two corresponding 3D line endpoints $\{P_1,P_2\}$ as follows:
\begin{equation}
\label{eq:line_err}
r_{l'}  = l'\begin{bmatrix} \pi(\widehat{R}P_1+\hat{t}\,) & \pi(\widehat{R}P_2+\hat{t}\,) \\ 1 & 1\end{bmatrix}
\end{equation}
Additionally, the 2D-to-2D point correspondences without depth can also be taken into account by using the residual introduced in \cite{DEMO}:

\begin{equation}
\label{eq:rep_err}
r_{p,p'}  = \lambda \begin{bmatrix} \hat{t}_y-\hat{t}_z y'& \hat{t}_z x'-\hat{t}_x & \hat{t}_x y'-\hat{t}_y x'\end{bmatrix} \widehat{R}\dot{p}
\end{equation}

Here, we use a factor $\lambda$ to down-weight these residuals, whereas in \cite{DEMO}, the reprojection errors of the 3D-to-2D matches are in fact scaled by the depth, therefore their impact on the overall optimization depends on the scene depth and metric unit. The value of $\lambda$ is set heuristically to 0.01.

These heterogeneous residuals are then stacked together and minimized by using a non-linear least squares algorithm (i.e. Levenberg-Marquart). Formally, let $S_1$ be a set of 3D-to-2D point matches; $S_2$, a set of 2D-to-3D point matches; $S_3$, a set of 3D-to-2D line matches; $S_4$, a set of 2D-to-3D line matches: and $S_5$, a set of 2D-to-2D point matches, then the optimal pose is  given by the minimization of the following joint cost function:
\begin{equation}
\label{eq:tukey}
\begin{aligned}
E(\xi)= & \sum_{p\in S_1}r_p^\top W_p r_p + \sum_{p'\in S_2}r_{p'}^\top W_{p'}  r_{p'} + \sum_{l\in S_3} r_{l} W^{}_{l}  r_{l}^\top  \\  + & \sum_{l'\in S_4}r_{l'} W^{}_{l'} r_{l'}^\top +\sum_{p,p'\in S_5} w^{}_{p,p'}  r^2_{p,p'}
\end{aligned}
\end{equation}
where $W$ are diagonal matrices containing the Tukey weights $w$, which are computed separately per residual-type in an iteratively re-weighted least-squares fashion. \par
Degenerate feature configurations are addressed by imposing a minimum of 3 total matches with depth and by checking, after the optimization, the uncertainty of the pose parameters $\xi$, which is approximated as the inverse of the Gauss-Newton approximation to the Hessian, based on the squared Mahalanobis distance (also known as the uncertainty back-propagation \cite{hartley2003multiple}):
\begin{equation}
\label{eq:pose_uncertainty}
\Sigma_{\xi} = (J_r^\top \Sigma_{r}^{-1} J_r)^{-1}
\end{equation}
where $J_r$ is the combined Jacobian matrix of the residual functions with respect to the estimated pose parameters, evaluated at the solution, and $\Sigma_{r}$ is a diagonal matrix containing the uncertainties of the residuals. Although $\Sigma_{r}$ can be derived by propagating the uncertainty of the 3D feature coordinates, in this work, for simplicity, we assume homoscedasticity by letting the uncertainty of a residual be the variance of the residuals of the same type.

\par If the largest eigenvalue of the 3$\times$3 block of $\Sigma_{\xi}$ corresponding to the translation exceeds a fixed threshold then the configuration is considered degenerate and a decaying velocity model is applied instead to estimate the pose.

\subsection{Point and Line Image Features}
At every frame, point and line features are extracted from the RGB image. Specifically for points, we make use of SURF features, whereas for lines, we use the LSD \cite{LSD} method to detect line segments and then extract binarized LBD \cite{LBD} descriptors (implemented in OpenCV) from the detected line segments. Features correspondences are then established by matching the descriptors between consecutive frames. Small frame-to-frame motion is assumed to prune away matches which are geometrically far: After performing a $k$-NN descriptor search for a given point $p$, we select the closest match in $k$ whose point coordinates lie in a circular region defining the neighbourhood of $p$, whereas, for a given line, we accept the closest line match in $k$ that has a similar slope angle and distance from origin, according to their line Hessian normal equations.

\subsection{Point Depth Sampling and 3D Line Fitting}
\label{sec:depth_sampling}
For a calibrated RGB-D sensor, the 3D location of feature points is directly obtained from the depth map by back-projecting the corresponding 2D coordinates. \par However, a 2D line segment contains multiple depth samples across its length, which suggests that one should exploit them instead of simply back-projecting the two depth pixels corresponding to the line endpoints as these are noisy, may be missing or correspond instead to the background, for lines tend to be detected on the object contours where depth is discontinuous. Hence, to obtain the 3D lines we propose to use the robust method of \cite{PLVO} with two simple modifications to address the computational cost of the original framework:

\begin{enumerate}
\item Using the Euclidean point-to-line distance within the RANSAC outlier filtering instead of the originally proposed Mahalanobis distance, for the two metric have not shown significant differences in terms of pose estimation accuracy, in our experiments (Section \ref{sec:experiments}),  however the computation of the Mahalanobis distance requires either inverting the uncertainty matrices of the line depth samples or applying whitening transformations to them \cite{stereoLines08}. In our implementation, using Eigen library, computing the Mahalanobis distances makes the process coarsely 3 times slower, thus we resort to use simply the Euclidean distance with an inlier threshold of 3 cm. 
	\item The final line fitting step in \cite{PLVO} is cast as a non-linear least squares problem in order to incorporate the line uncertainty. Instead, we perform PCA on the consensus set of line samples: $X$, given by RANSAC. Let the covariance of $X$ be factorized as: $\Sigma_X=VSV^\top$, then we parameterize the line, according to the spread of samples, with the two 3D points: $\bar{P}\pm\sigma_1 V_1$, where $\bar{P}$ is the mean of the samples, $\sigma_1$ is the square root of the largest eigenvalue in $S$ and $V_1$ is the corresponding eigenvector.
\end{enumerate}

\begin{figure}[thpb]
\centering
	\begin{tabular}{@{}c@{}c@{}}
		\includegraphics[scale=0.6]{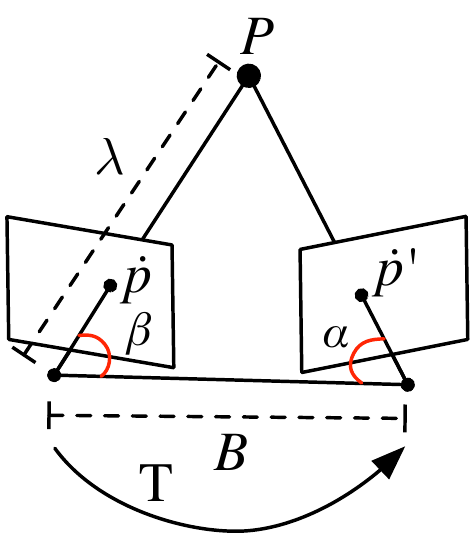} & \quad
		\includegraphics[scale=0.6]{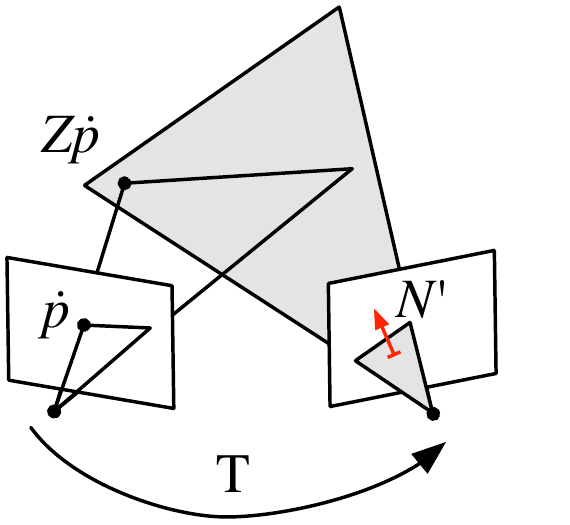} \\
		(a) & (b)
	\end{tabular} 
	\caption{(a) Ideal point triangulation for point matches that satisfy the epipolar constraint (b) Line triangulation as a line-to-plane intersection.}
	\label{fig4}
	\vspace*{-2mm} 
\end{figure}

\subsection{Line Depth Estimation}
As depicted in Fig. \hyperref[fig4]{\ref*{fig4}.b}, given a pair of line segment matches, the depth of one line segment endpoint $p$ can be derived from the point-to-plane distance:

\begin{equation}
\label{eq:line_depth_est}
N'(R Z \dot{p}+t) = 0 \Leftrightarrow Z = -\frac{N' t}{N' R \dot{p}}
\end{equation}
where $N'$ is the normal of the projective plane obtained by the cross-product of the two endpoints on the second image. This expression is ill-posed when the rotated point lies on the projective plane of the other frame, therefore it may result in a negative $Z$ even for a correct line match. Thus, to detect outliers, endpoints with invalid depth estimates are assigned arbitrary positive depth and then reprojected on the second frame to check the point-to-line distance.

\subsection{Point Depth Estimation}

In practice, the feature point matches, resulting from the descriptor matching, do not meet exactly the epipolar constraint, meaning that their projection rays do not intersect.
This issue is addressed by employing the commonly used linear triangulation method (described in \cite{hartley2003multiple}, p. 312) to estimate the 3D point coordinates, which in turn we use to obtain the corrected projected point coordinates.
The projection errors between the original and the corrected coordinates are checked to detect outliers. Additionally, for invalid depth estimates we use an outlier detection procedure equivalent to the one described in the previous section. Since the problem of point triangulation is overdetermined, the triangulation method \cite{hartley2003multiple} relies on a least-squares solution, which makes the derivation of depth uncertainty less straightforward, however, given the corrected coordinates we can now arrive, through the trigonometry represented in Fig. \hyperref[fig4]{\ref*{fig4}.a}, at the following formula:

\begin{equation}
\label{eq:point_depth_est}
Z = \frac{\lambda}{\lVert\dot{p}\rVert }  \qquad \lambda= \frac{B}{\sin{\beta}\tan{\alpha}+\cos{\beta}}
\end{equation}
where:
\begin{equation}
\label{eq:point_depth_est_cont}
\cos{\beta} = \frac{-t R \dot{p}}{\lVert\dot{p}\rVert B} \qquad \cos{\alpha} = \frac{t\dot{p}'}{\lVert\dot{p}'\rVert B}
\end{equation}

\subsection{Uncertainty of Depth and Pose Estimates}
The uncertainty of the depth estimation is modelled in terms of inverse depth variance to allow initialization of 3D features that are far away from the camera.
Based on the first order error propagation, the uncertainty of the general inverse depth denoted here as: $\rho=Z^{-1}$ is given by:
\begin{equation}
\label{eq:uncertainty_prop}
\sigma^2_{\rho} = J_{\rho}\Sigma J_{\rho}^\top
\end{equation}
where $J_{\rho}$ is the Jacobian of the inverse depth function with respect to the input variables and $\Sigma$ is the variables' uncertainty matrix. Specifically, this inverse depth function corresponds to the inverse of (\ref{eq:line_depth_est}) and (\ref{eq:point_depth_est}) for, respectively, a line endpoint and a point. \par
Regarding the variables' uncertainties, depth estimation for a point correspondence depends on two image points: $\{u_1,v_1,u_1',v_1'\}$ and on the stereo pose parameters $\xi$, whereas the depth estimation of a line endpoint depends on three image points $\{u_1,v_1,u_2',v_2',u_3',v_3'\}$ and $\xi$. While the uncertainty of each point coordinate is set to 1 pixel, the $6\times6$ block of $\Sigma$ corresponding to the pose is derived from the propagation of the pose uncertainty as follows.\par Let the parameters $\xi^{(k|1)}$ express the camera pose at frame $1$ with respect to frame $k$ and $\Sigma^{(k+1|k)}_{\xi}$ be the uncertainty obtained by (\ref{eq:pose_uncertainty}), then $\xi^{(k+1|1)}$ is given by the state transition function:
\begin{equation}
\label{eq:pose_comp}
f(\xi^{(k|1)},\xi^{(k+1|k)})= \begin{bmatrix} R^{(k+1|k)} t^{(k|1)} + t^{(k+1|k)} \\ q^{(k+1|k)} \otimes q^{(k|1)}\end{bmatrix} 
\end{equation}
with $\otimes$ denoting the quaternion multiplication, and the respective pose uncertainty is given according to the EKF propagation equation as:
\begin{equation}
\label{eq:pose_var_prop}
\Sigma^{(k+1|1)}_{\xi} = F \Sigma^{(k|1)}_{\xi} F^\top + G\Sigma^{(k+1|k)}_{\xi}G^\top
\end{equation}
where $F$ and $G$ are the Jacobians of the first 6 columns of (\ref{eq:pose_comp}) with respect to $\xi^{(k|1)}$ and $\xi^{(k+1|k)}$, respectively.
\par
\begin{figure}
\centering
		\includegraphics[scale=0.6]{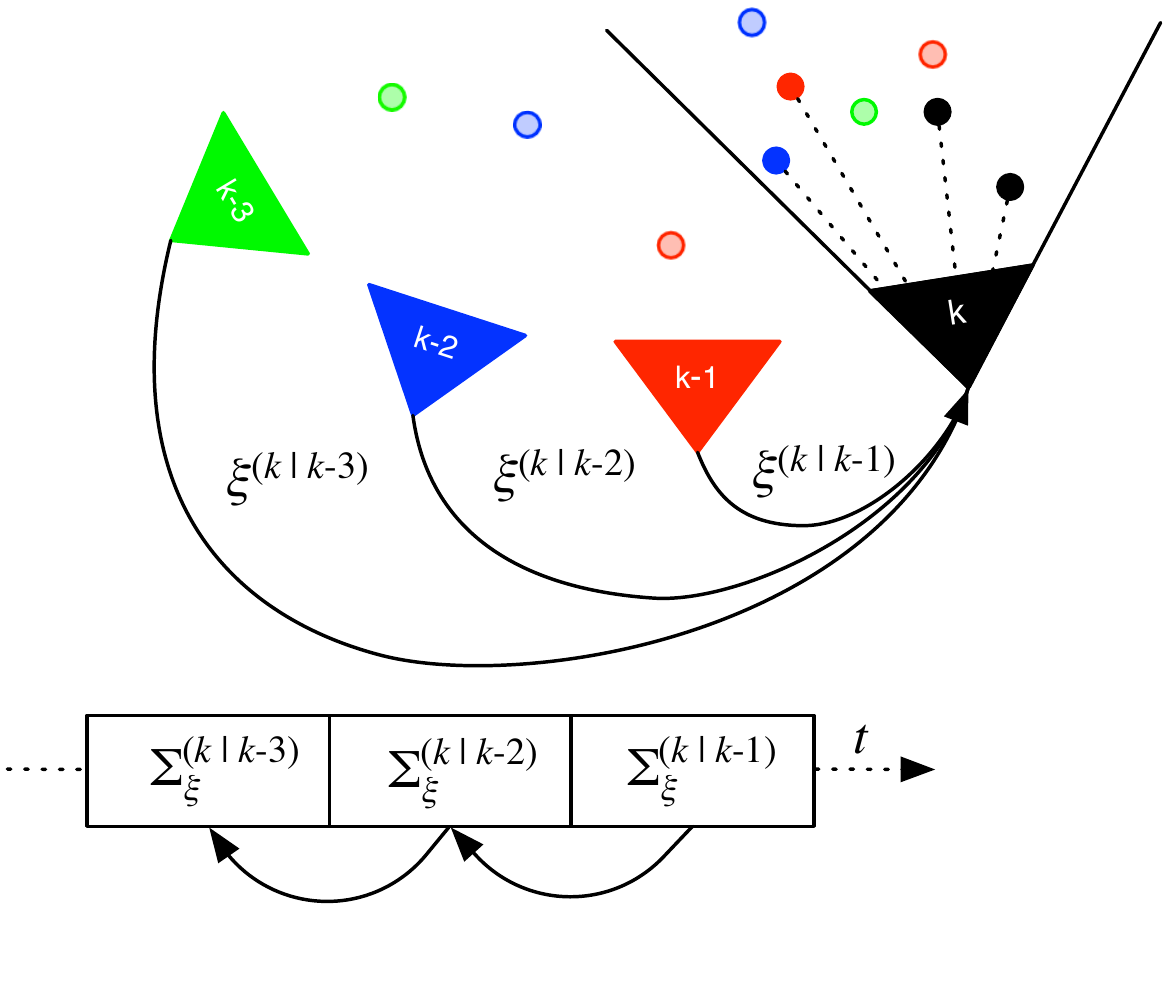}
		\vspace*{-5mm} 
	\caption{Illustration of the depth map registration at frame $k$ along with a sliding window of pose uncertainties, which is used both to limit the propagation of depth measurements and derive the uncertainty of the triangulated depth. In this example, the uncertainty of the transformation between $k$ and $k-3$ is considered to be too high therefore all depth measurements from frame $k-3$ are forgotten.}
	\label{fig5}
	\vspace*{-2mm} 
\end{figure}

To avoid the redundancy of calculating the stereo pose uncertainty for each feature, a sliding window of transformations to the current frame and the respective uncertainties is maintained and updated using (\ref{eq:pose_comp})  and (\ref{eq:pose_var_prop}), as depicted in Fig. \ref{fig5}.

\subsection{Fusion of Depth Estimates}
Sequential filtering is performed in order to fuse the estimates from multiple views. 
Each point, being tracked, has associated a 1D Gaussian PDF to represent its inverse depth state estimate. This is initialized as $\mathcal{N}(0,1)$ (with infinite depth) and it is updated whenever there is a new triangulation.
Given the PDF of a new triangulated inverse depth estimate, represented as: $\mathcal{N}(\rho_2, \sigma_{2}^2)$, the prior depth estimate state:  $\mathcal{N}(\rho_1, \sigma_{1}^2)$  is updated by the product of two Gaussians:
\begin{equation}
\label{eq:depth_fusion}
\rho_{fused} = \frac{\rho_1 \sigma^2_2 +\rho_2 \sigma^2_1 }{\sigma^2_{1} +\sigma^2_{2}}            \qquad \sigma^2_{fused}=\frac{\sigma^2_{1} \sigma^2_{2}}{\sigma^2_{1} +\sigma^2_{2}}
\end{equation}

For point features, when $\sigma^2_{fused}$ drops below a fixed threshold, we use $\rho_{fused}$ to back-project a 3D point, so that it can be used in the next frame. For line features, the 3D endpoints are initialized only when both uncertainties are sufficiently low.

\subsection{Depth Map Registration}
\label{sec:depth_registration}
Depth measurements from past frames are propagated as a registered point cloud to recover the missing depth measurements of the current depth map. A denser depth map can then be obtained by transforming and projecting the registered point cloud to the current camera view. At each frame, the point cloud is filtered and updated with new measurements, as shown in Fig. \ref{fig5}, according to the following criteria:

\begin{itemize}
	\item Transformation uncertainty: To avoid inaccurate point registration due to the growing pose error, points measured at frames whose transformation to the current frame has an uncertainty indicator higher than a certain limit are removed from the point cloud. Specifically, a sliding window of transformation uncertainties is maintained and their largest eigenvalues are used as an uncertainty indicator, as described in Section \ref{sec:pose_est}.
	\item Field-of-view: Only points within the FOV of the color camera are maintained.
	\item One point per pixel: Each point must have a unique image projection pixel. In case of conflict, the most recent point is kept.
\end{itemize}
While one could devise more sophisticated methods for combining the depth measurements, e.g., averaging depth observations, an important advantage of this simple method is the efficiency, since the size of the point cloud is bounded by the image resolution, due to the last two criteria. 

\section{Experiments}
\label{sec:experiments}
The performance of our visual odometry method is evaluated on the TUM RGB-D dataset \cite{tum_dataset} and on an author-collected RGB-D dataset. All sequences have been captured with a structured-light camera and a color camera operating at 30 fps with a resolution of 640$\times$480.

\setlength{\tabcolsep}{3pt}
\begin{table}
	\caption{RMSE of relative pose per second for line odometry with three different 3D line fitting schemes and no depth estimates. Direct sampling simply back-projects the 2D line endpoints whereas the others use the RANSAC method described in Section \ref{sec:depth_sampling} with different types of inlier distances.} 
\centering   
\begin{tabular}{l@{ }ccc}
\hline\\[-1ex] 
Sequence & Direct sampling & Euclidean dist. & Mahalanobis dist.\\
[0.5ex] 
\hline\\[-1ex] 
fr1/desk & 42 mm / 2.2\degree  & 41 mm / 2.2\degree & 40 mm / 2.2\degree \\[0.2ex] 
fr2/large\_no\_loop  &  85 mm / 1.2\degree & 74 mm / 1.2\degree & 75 mm / 1.2\degree \\[0.5ex] 
\hline \\
\end{tabular}
	\label{tab:line_sampling}
\end{table}

{\renewcommand{\arraystretch}{1.2}
\begin{table}[]
\centering
\caption{RMSE of relative pose per second on TUM sequences. We report also, to our knowledge, the best translational errors obtained by other visual odometry methods. The symbol $\dagger$ marks sequences captured in small environments where depth measurements are abundant, thus including the depth estimates did not change significantly the performance.} 
\label{tab:TUM_results}
\begin{tabular}{|l|c|c|c|c||c|}
\hline
\multirow{2}{*}{Sequence} & \multirow{2}{*}{\begin{tabular}[c]{@{}c@{}}Depth \\ estimates\end{tabular}} & \multicolumn{3}{c||}{SPLODE (Mode A)} & \multirow{2}{*}{\begin{tabular}[c]{@{}c@{}}Best\\ (others)\end{tabular}} \\ \cline{3-5}
 &  & Points & Lines & All &  \\ \hline
fr1/desk $\dagger$ & No & \begin{tabular}[c]{@{}c@{}}32 mm\\ 2.4\degree\end{tabular} & \begin{tabular}[c]{@{}c@{}}41 mm\\ 2.2\degree\end{tabular} & \begin{tabular}[c]{@{}c@{}}29 mm\\ 2.2\degree\end{tabular} & \begin{tabular}[c]{@{}c@{}}26 mm \\ {\cite{gutierrez2016dense}}\end{tabular} \\ \hline
fr1/360 $\dagger$ & No & \begin{tabular}[c]{@{}c@{}}89 mm\\ 3.5\degree\end{tabular} & \begin{tabular}[c]{@{}c@{}}80 mm\\ 2.6\degree\end{tabular} & \begin{tabular}[c]{@{}c@{}}66 mm\\ 3.0\degree\end{tabular} & \begin{tabular}[c]{@{}c@{}}84mm\\ {\cite{PLVO}}\end{tabular} \\ \hline
\multirow{2}{*}{\begin{tabular}[c]{@{}l@{}}fr2/\\ 360\_hemisphere\end{tabular}} & No & \begin{tabular}[c]{@{}c@{}}78 mm \\ 1.6\degree\end{tabular} & \begin{tabular}[c]{@{}c@{}}90 mm\\ 1.1\degree\end{tabular} & \begin{tabular}[c]{@{}c@{}}74 mm\\ 1.0\degree\end{tabular} & \multirow{2}{*}{-} \\ \cline{2-5}
 & Yes & \begin{tabular}[c]{@{}c@{}}70 mm\\ 1.0\degree\end{tabular} & \begin{tabular}[c]{@{}c@{}}77 mm\\ 1.0\degree\end{tabular} & \begin{tabular}[c]{@{}c@{}}66 mm\\ 1.0\degree\end{tabular} &  \\ \hline
\multirow{2}{*}{\begin{tabular}[c]{@{}l@{}}fr2/\\ large\_no\_loop\end{tabular}} & No & \begin{tabular}[c]{@{}c@{}}91 mm\\ 1.2\degree\end{tabular} & \begin{tabular}[c]{@{}c@{}}74 mm\\ 1.2\degree\end{tabular} & \begin{tabular}[c]{@{}c@{}}74 mm\\ 1.1\degree\end{tabular} & \multirow{2}{*}{\begin{tabular}[c]{@{}c@{}}96 mm\\ {\cite{whelan2013fusion}}\end{tabular}} \\ \cline{2-5}
 & Yes & \begin{tabular}[c]{@{}c@{}}89 mm\\ 1.2\degree\end{tabular} & \begin{tabular}[c]{@{}c@{}}65 mm\\ 1.1\degree\end{tabular} & \begin{tabular}[c]{@{}c@{}}72 mm \\ 1.1\degree\end{tabular} &  \\ \hline
\multirow{2}{*}{\begin{tabular}[c]{@{}l@{}}fr2/\\ pioneer\_360\end{tabular}} & No & \begin{tabular}[c]{@{}c@{}}96 mm\\ 2.1\degree\end{tabular} & \begin{tabular}[c]{@{}c@{}}84 mm\\ 2.5\degree\end{tabular} & \begin{tabular}[c]{@{}c@{}}57 mm\\ 2.4\degree\end{tabular} & \multirow{2}{*}{-} \\ \cline{2-5}
 & Yes & \begin{tabular}[c]{@{}c@{}}67 mm\\ 1.9\degree\end{tabular} & \begin{tabular}[c]{@{}c@{}}72 mm\\ 2.5\degree\end{tabular} & \begin{tabular}[c]{@{}c@{}}53 mm\\ 1.9\degree\end{tabular} &  \\ \hline
\end{tabular}
\end{table}
}

\subsection{TUM RGB-D dataset}

We have conducted experiments on two distinct types of environments that are captured by the TUM RGB-D dataset: A small office and a large industrial hall denoted respectively as fr1 and fr2. While the sequences recorded in the former are commonly used for benchmarking RGB-D odometry and SLAM methods, few works have addressed some of the fr2 sequences. We believe this is mostly due to the large scene depth (see Fig. \ref{fig8}) which leads to insufficient depth measurements, which in turn causes tracking failures. Consequently, only fractions of these sequences were indeed evaluated in \cite{pinpointSLAM}. On the contrary, our results are reported for the full sequences, except for the fr2/large\_no\_loop which only has ground truth for about 20 \% of the sequence duration. As described in Section \ref{sec:pose_est}, tracking or pose estimation failures are handled by using a velocity model. \par
Performance was evaluated in terms of pose drift as the relative pose error (RPE) per second. First, Table \ref{tab:line_sampling} shows how the choice of 3D line fitting affects the pose error. We have observed that when the depth information is rich, simple back-projection of the endpoints seems enough, however when the depth measurements are severely compromised, the robust line fitting becomes necessary. No significant differences have been observed however between using the Mahalanobis or the Euclidean distance. \par 
As can be seen in Table \ref{tab:TUM_results}, including depth estimates reduces overall the odometry drift on the fr2 scene. Moreover, the number of frames, where tracking fails, is reduced, e.g., in the fr2/360\_hemisphere, while point odometry without depth estimates misses 29 frames, with the depth estimates, tracking is not interrupted. Since no significant differences were observed between the two depth estimation schemes in this dataset, we only report results for the mode A, which is preferred in terms of computational cost.

\begin{figure}[t]
\centering
	\begin{tabular}{@{}c@{}c@{}}
		\includegraphics[scale=0.31]{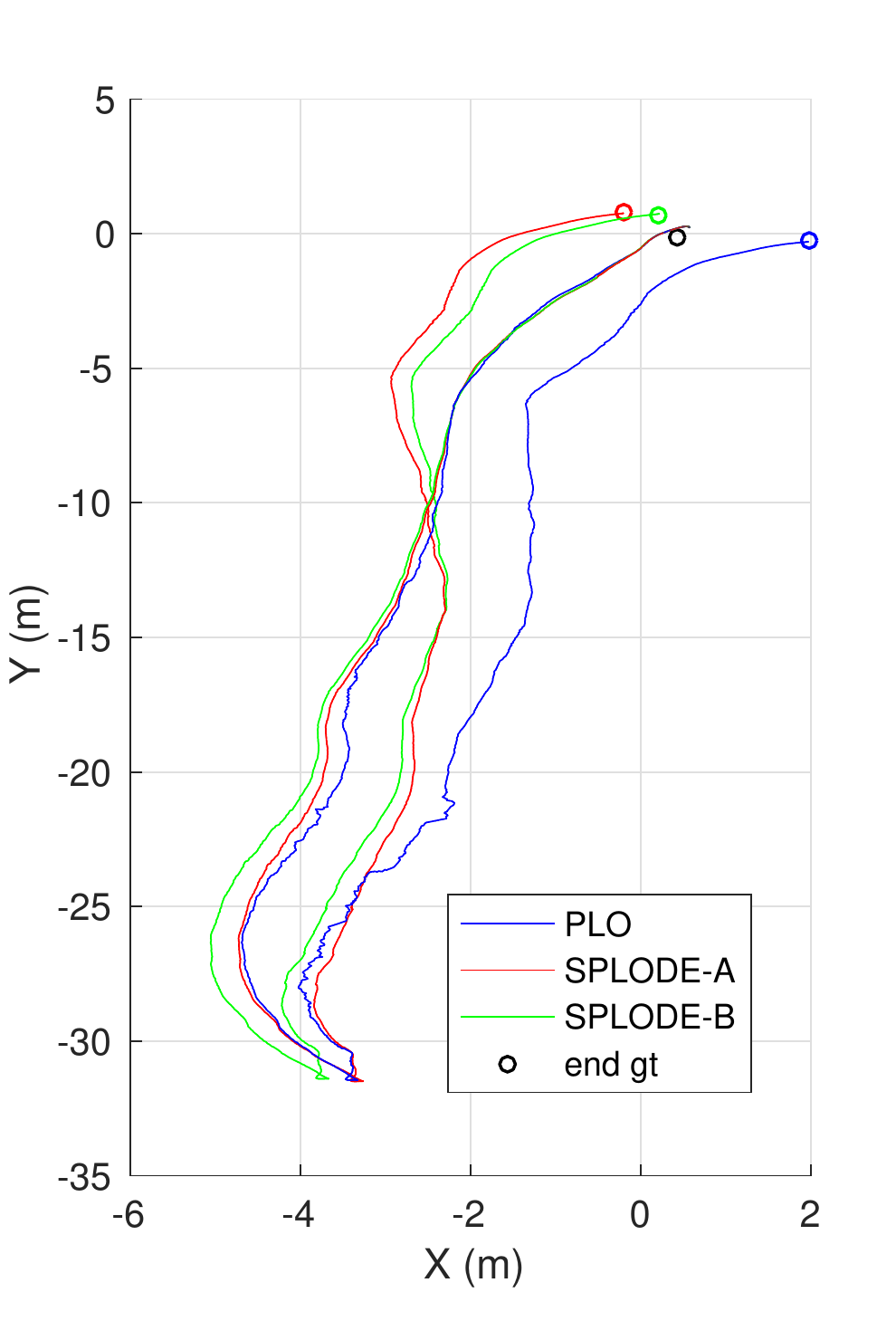} &
		\includegraphics[scale=0.31]{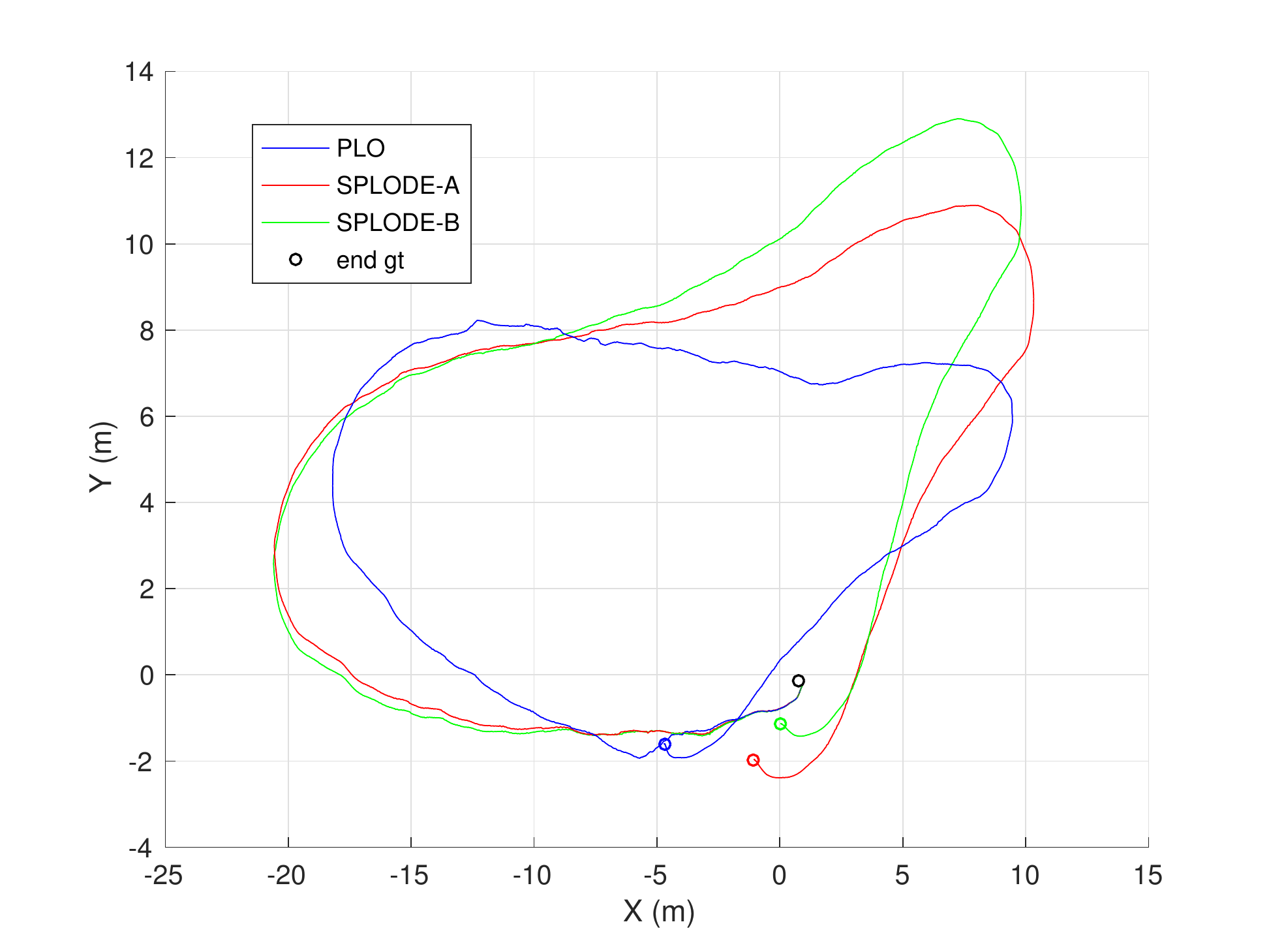} \\
	\end{tabular} 
		\vspace*{-3mm} 
	\caption{Top view of trajectories estimated by our point and line odometry without depth estimates: PLO, and with depth estimates: SPLODE-A and SPLODE-B for the respective schemes, on the foyer sequence (left image) and the park sequence (right image). The ground-truth of the last position is pinpointed as: end gt.}
	\label{fig6}
		\vspace*{-2mm} 
\end{figure}

\subsection{Author-Collected Dataset}

To further evaluate the robustness of the proposed method, we have captured two challenging RGB-D sequences, for depth sensors: A large foyer and a park under sun exposure, shown in Fig. \ref{fig1} and \ref{fig8}. The respective trajectories estimated by SPLODE are shown in Fig. \ref{fig6}, the final error of the trajectories is reported in Table \ref{tab:surrey_results} and a video is available at: \url{https://youtu.be/6lMwPCiCXZc}.
\par
As shown in Fig. \ref{fig6}, while pure RGB-D odometry with points and lines (i.e. PLO) fails, even with depth map registration, by making frequently gross errors or failing to estimate pose, the integration of depth estimates allows the method to perform well in both environments. In terms of features (see Table \ref{tab:surrey_results}), point-based odometry performs worst than line-based odometry in the foyer sequence due to the predominance of lines, on the other hand, point-based odometry performs well in the park sequence, whereas using just lines, as expected, is not sufficient, since straight lines are less common in nature.
\par
\begin{figure}[]
\centering
		\includegraphics[scale=0.31]{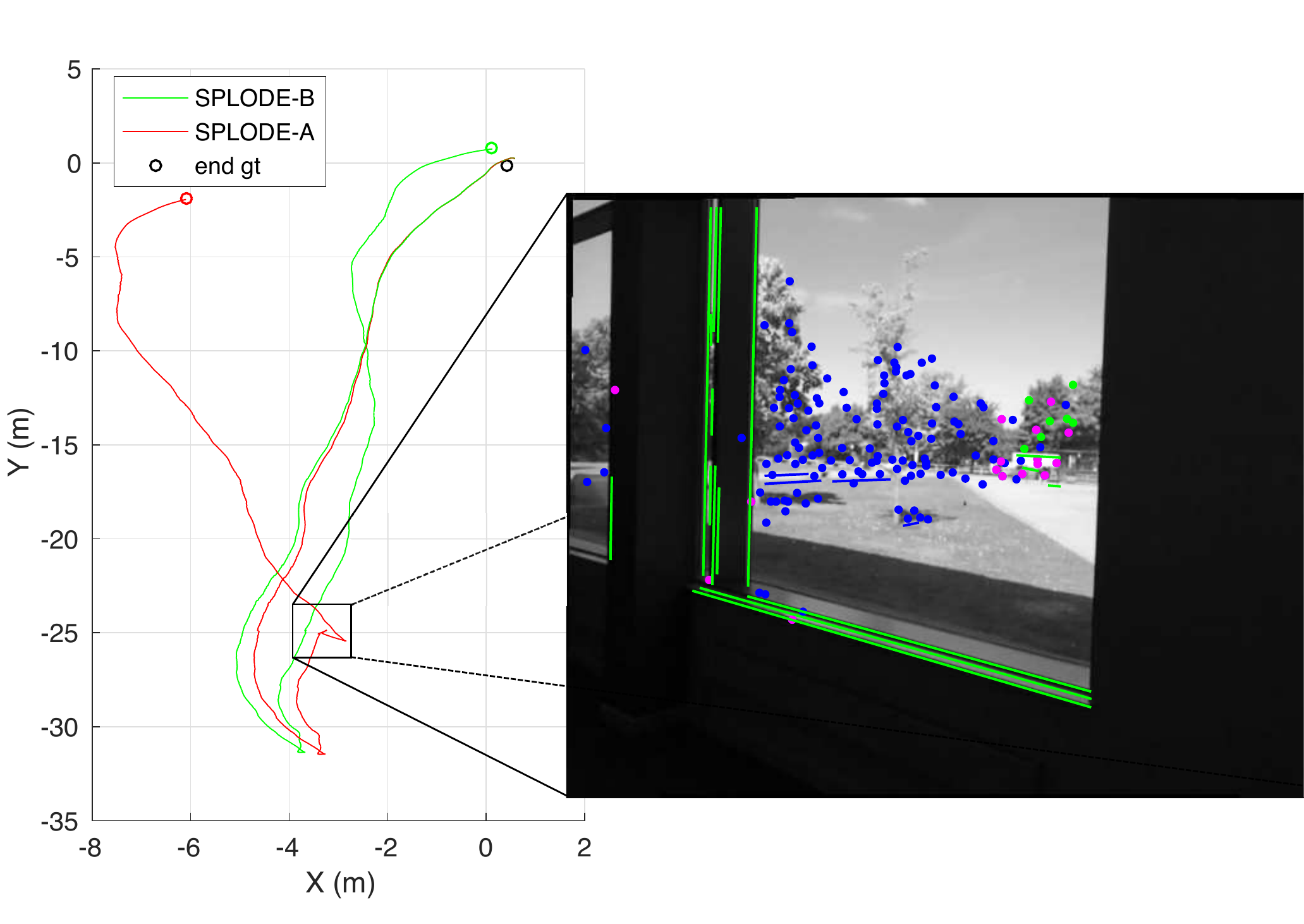}
		\vspace*{-1mm} 
	\caption{Trajectories obtained for the foyer sequence by disabling the uncertainty condition of the depth map registration. Features with two depth hypothesis (used by SLPODE-B) are shown in magenta and features with either depth measurements or estimates are shown respectively in green and blue. Spurious depth measurements, associated to several features observed through the window, are propagated by the depth map registration and eventually cause the mode A to fail, while the mode B is able to survive them.}
	\label{fig7}
	\vspace*{-2mm} 
\end{figure}
\begin{figure*}[t]
\centering
	\begin{tabular}{@{}c@{}c@{}c@{}c@{}}
		\includegraphics[scale=0.24]{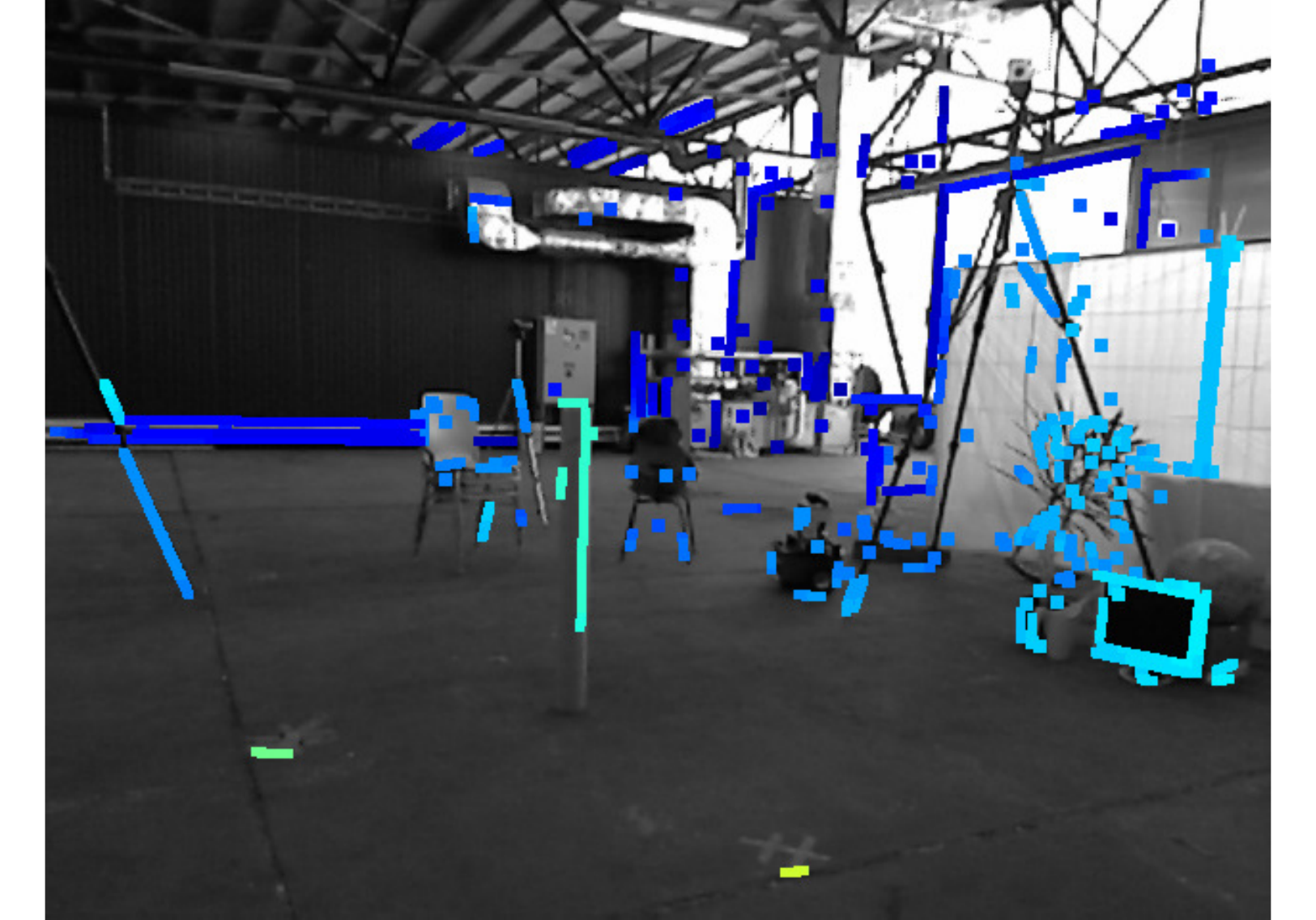} &
		\includegraphics[scale=0.24]{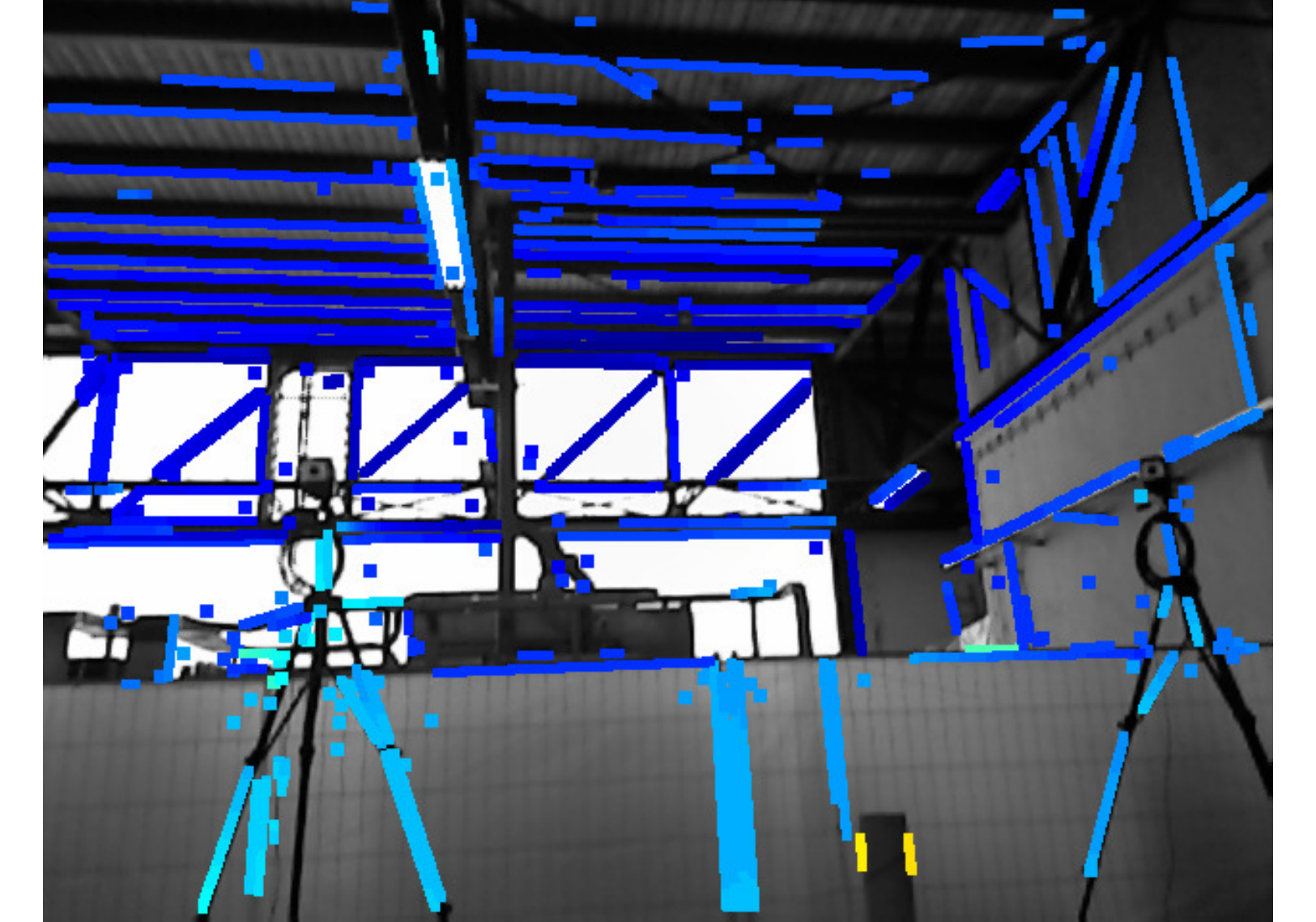} &
		\includegraphics[scale=0.24]{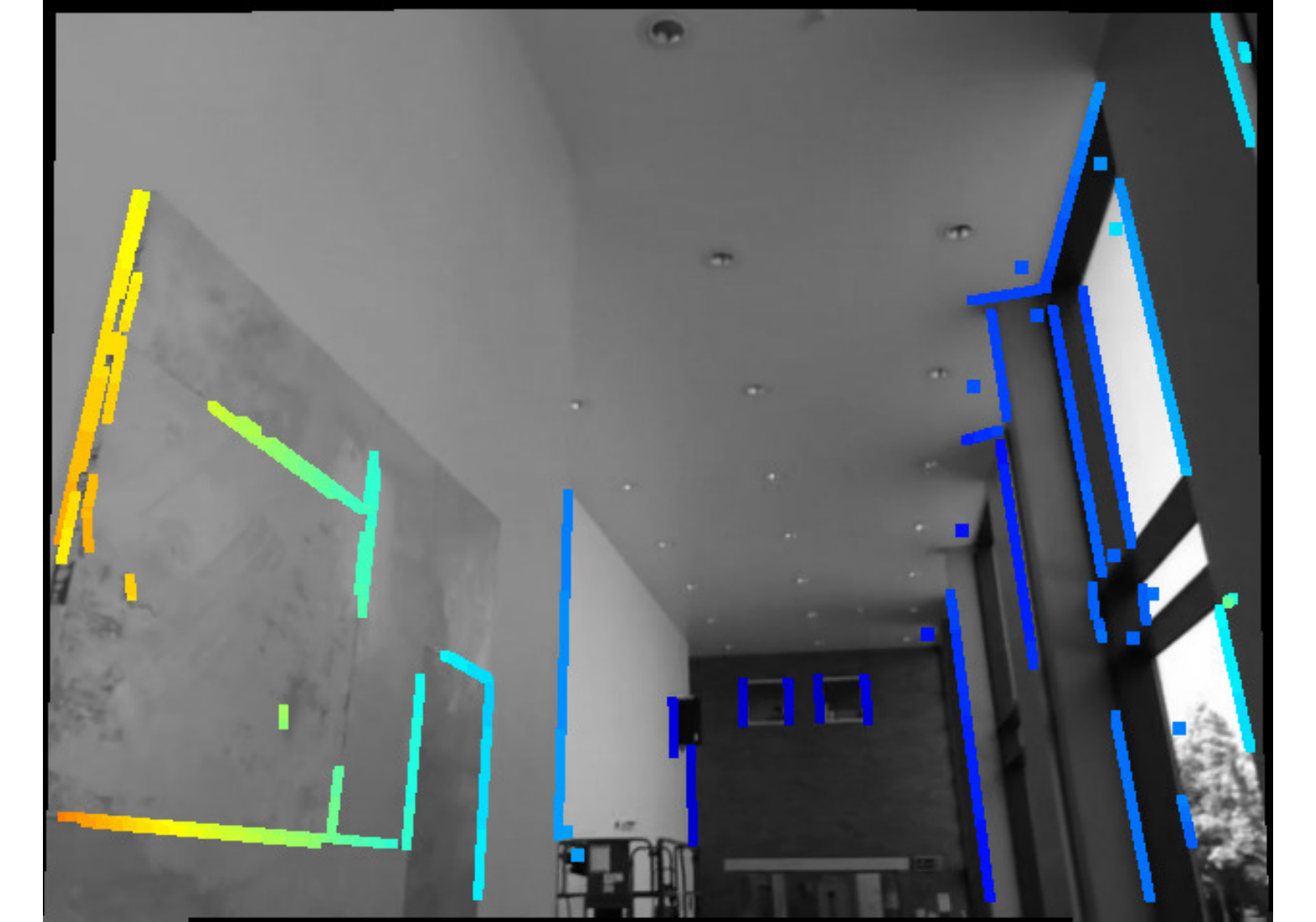} &
		\includegraphics[scale=0.24]{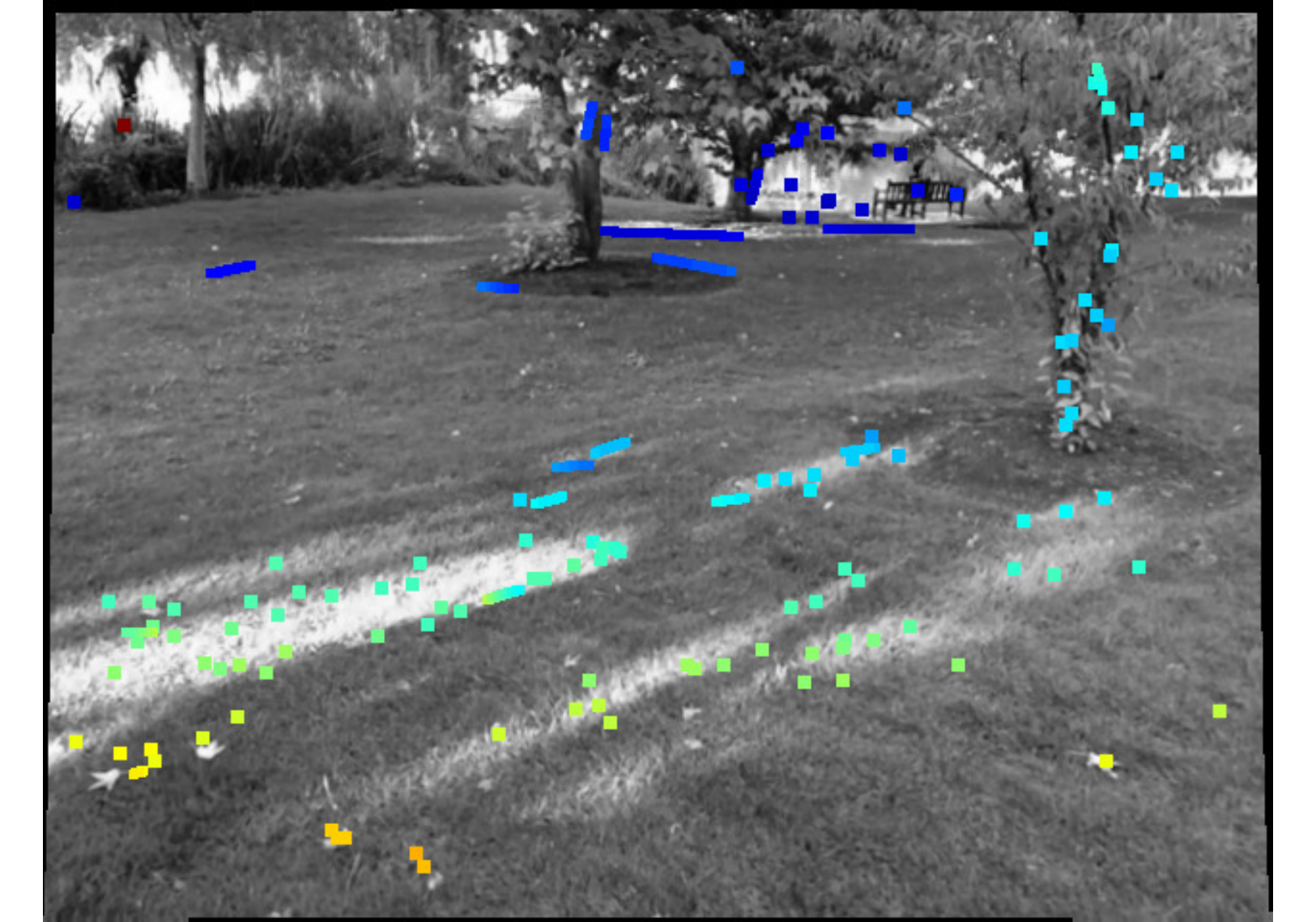} \\[-0.2ex] 
		\includegraphics[scale=0.24]{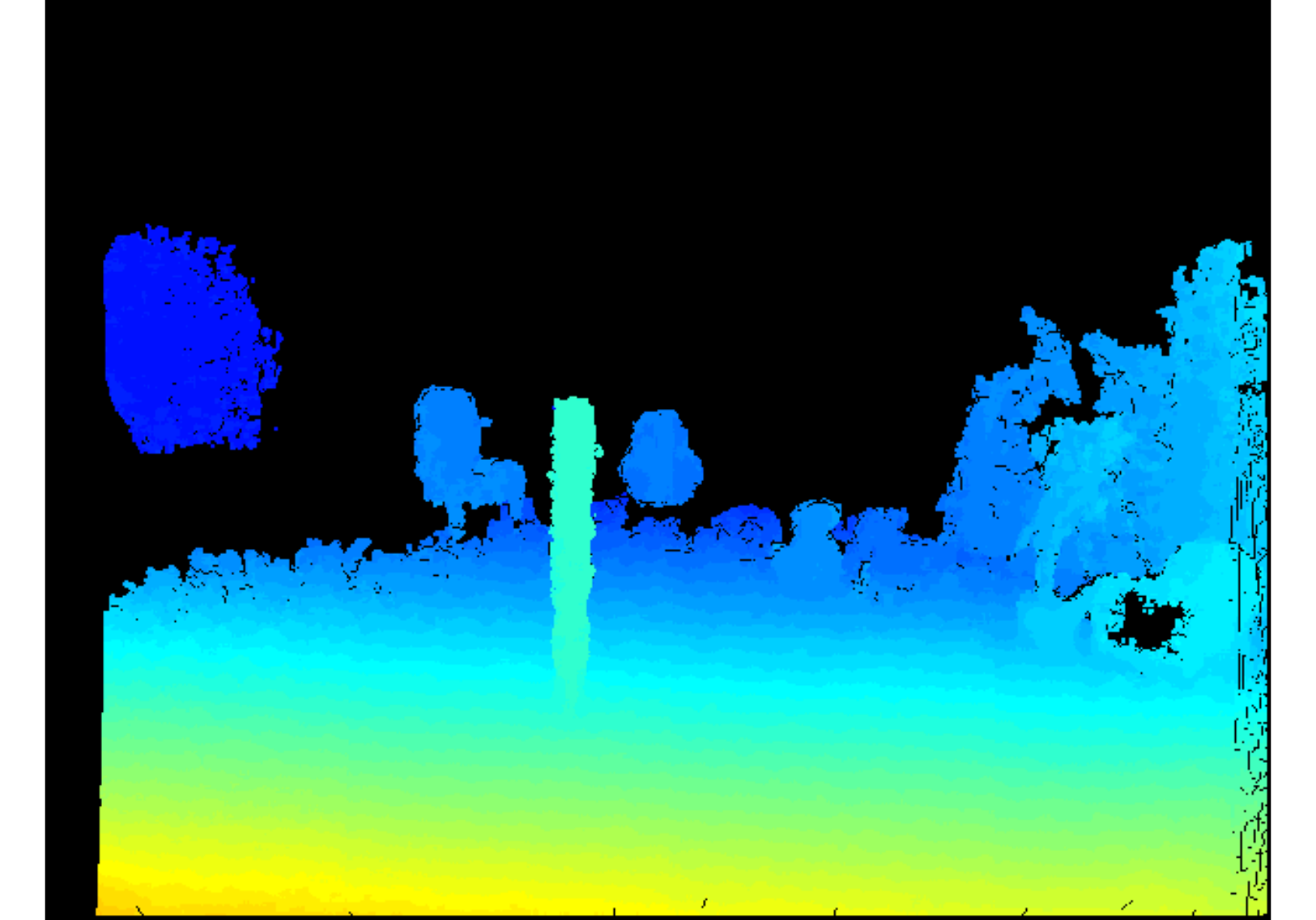} &
		\includegraphics[scale=0.24]{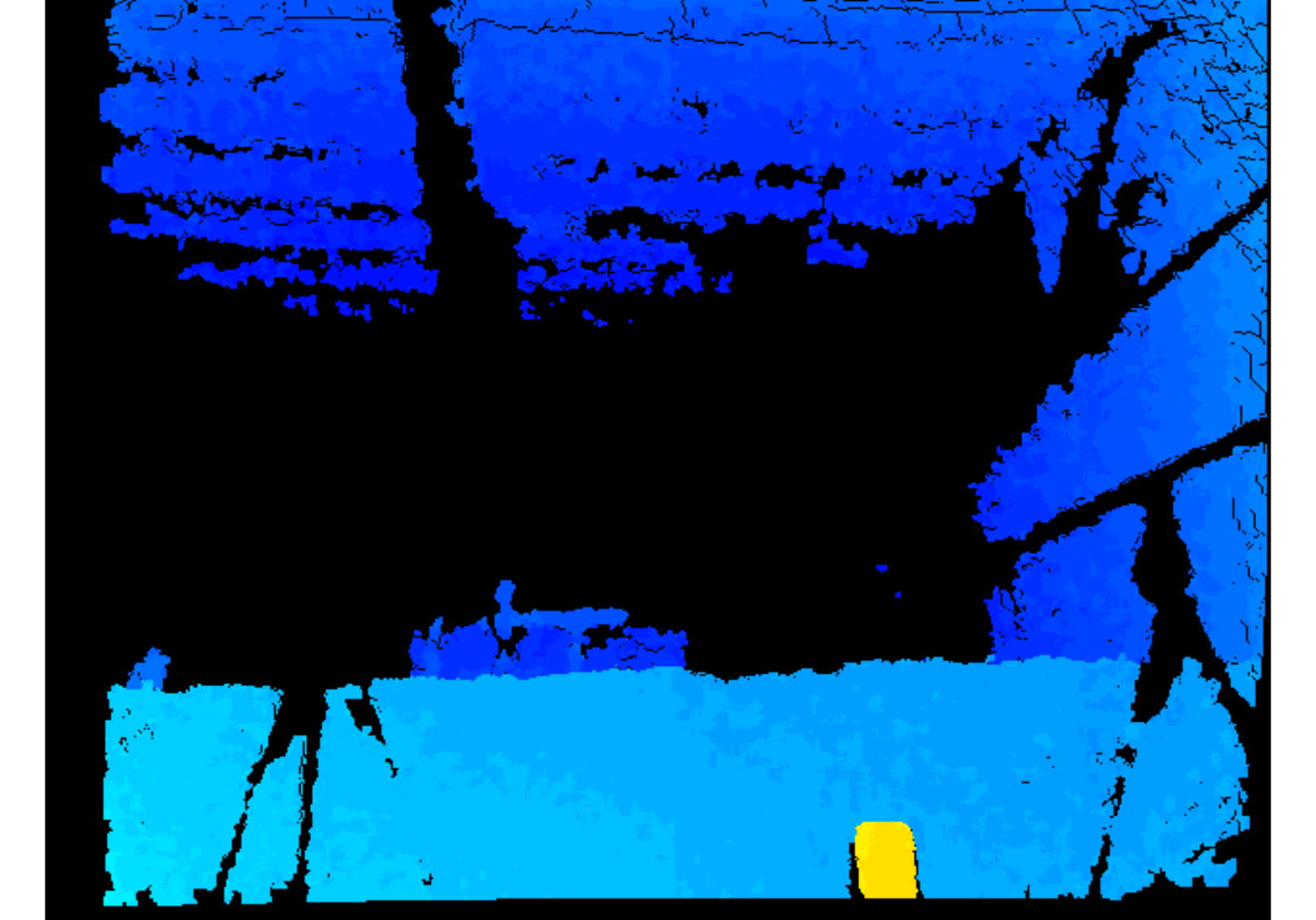} &
		\includegraphics[scale=0.24]{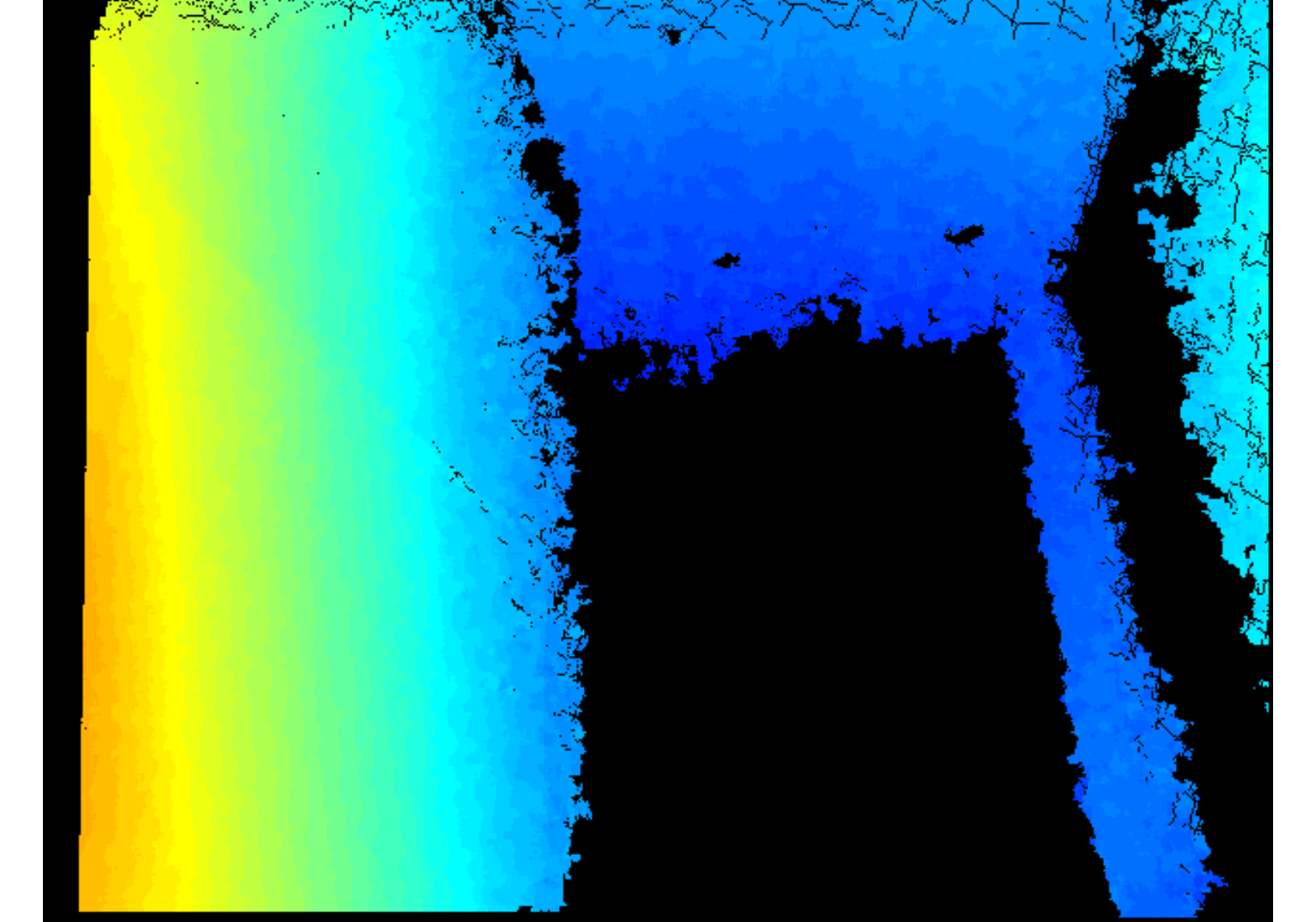} &
		\includegraphics[scale=0.24]{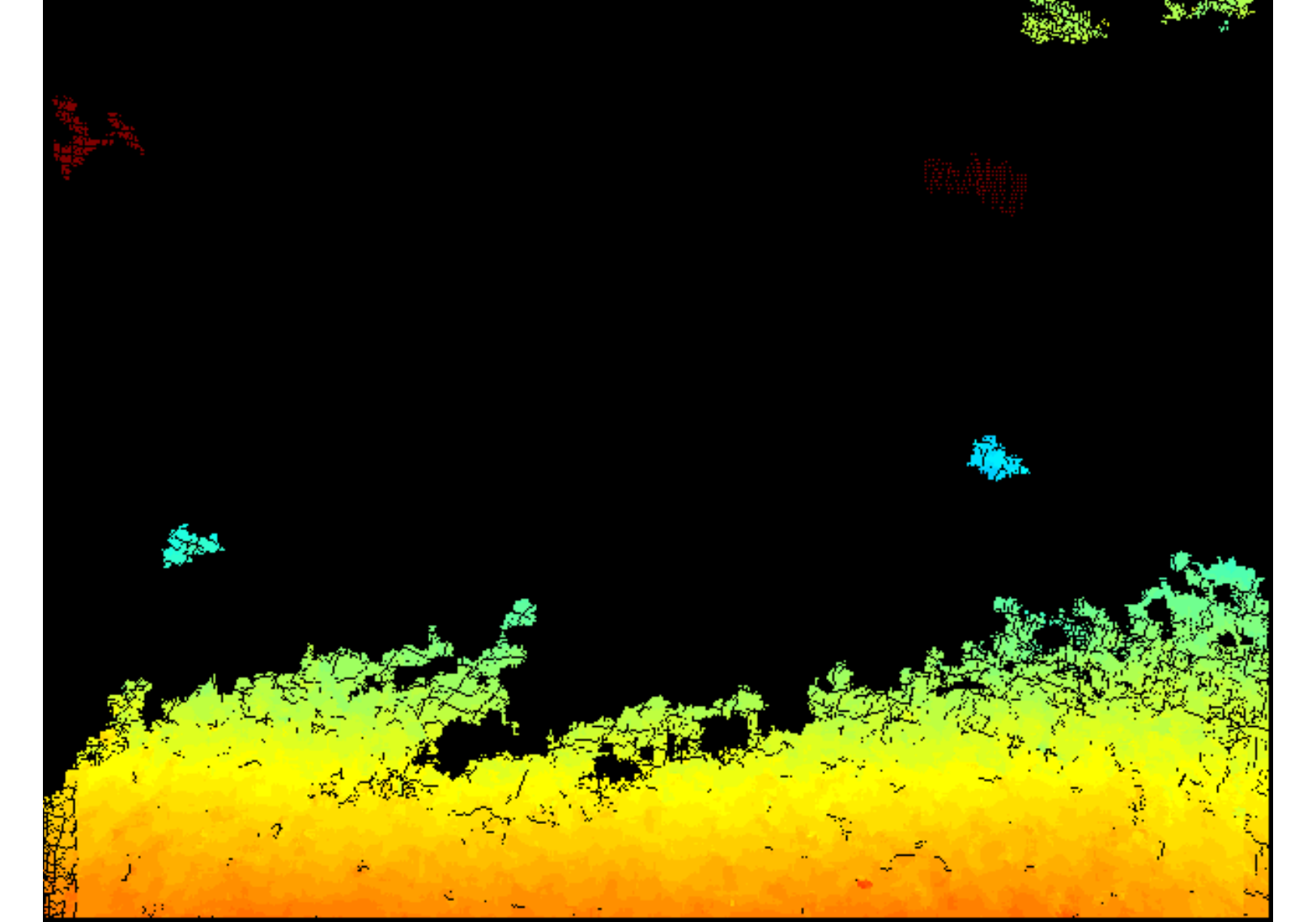} \\[-0.2ex] 
		\includegraphics[scale=0.183]{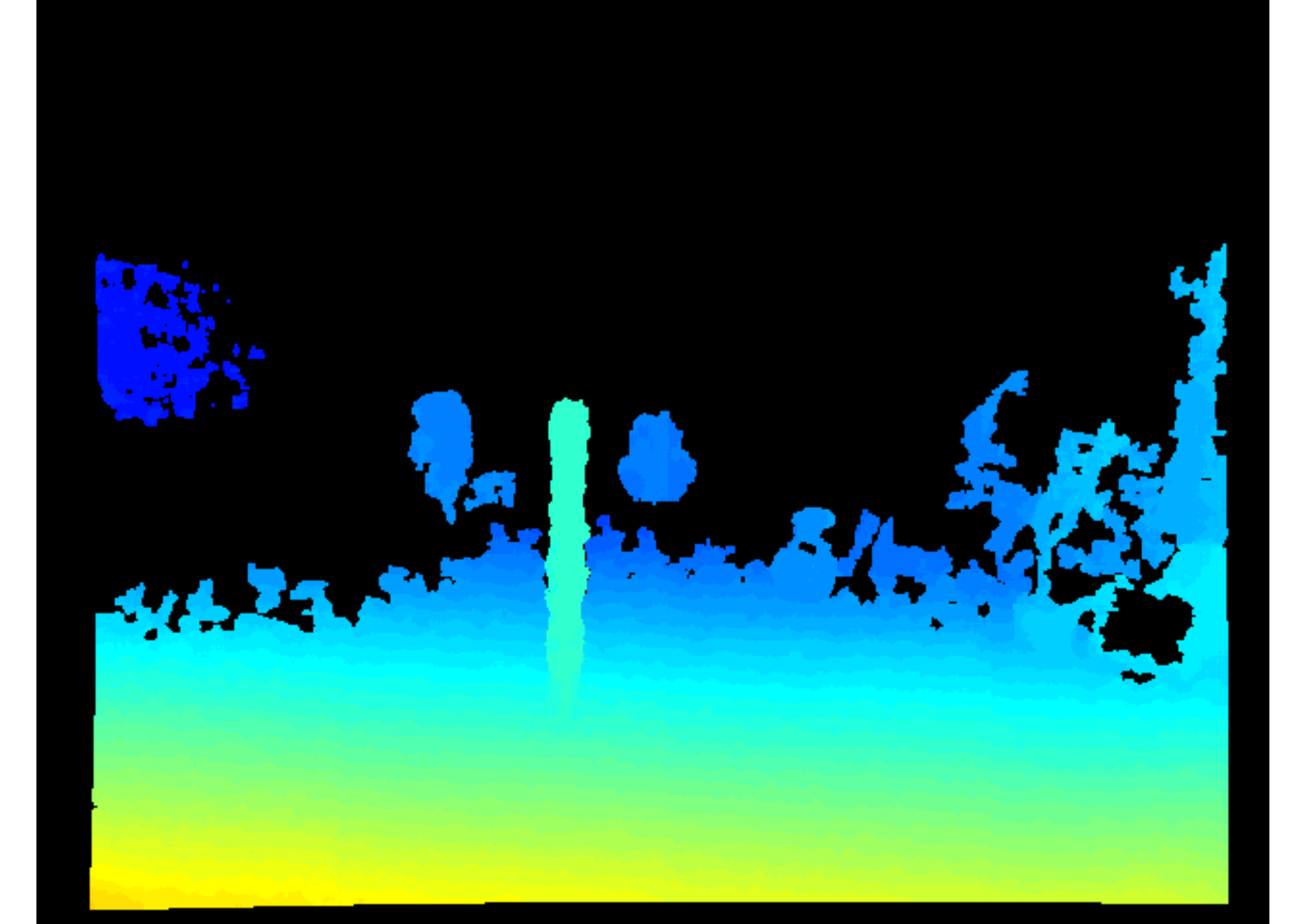} &
		\includegraphics[scale=0.183]{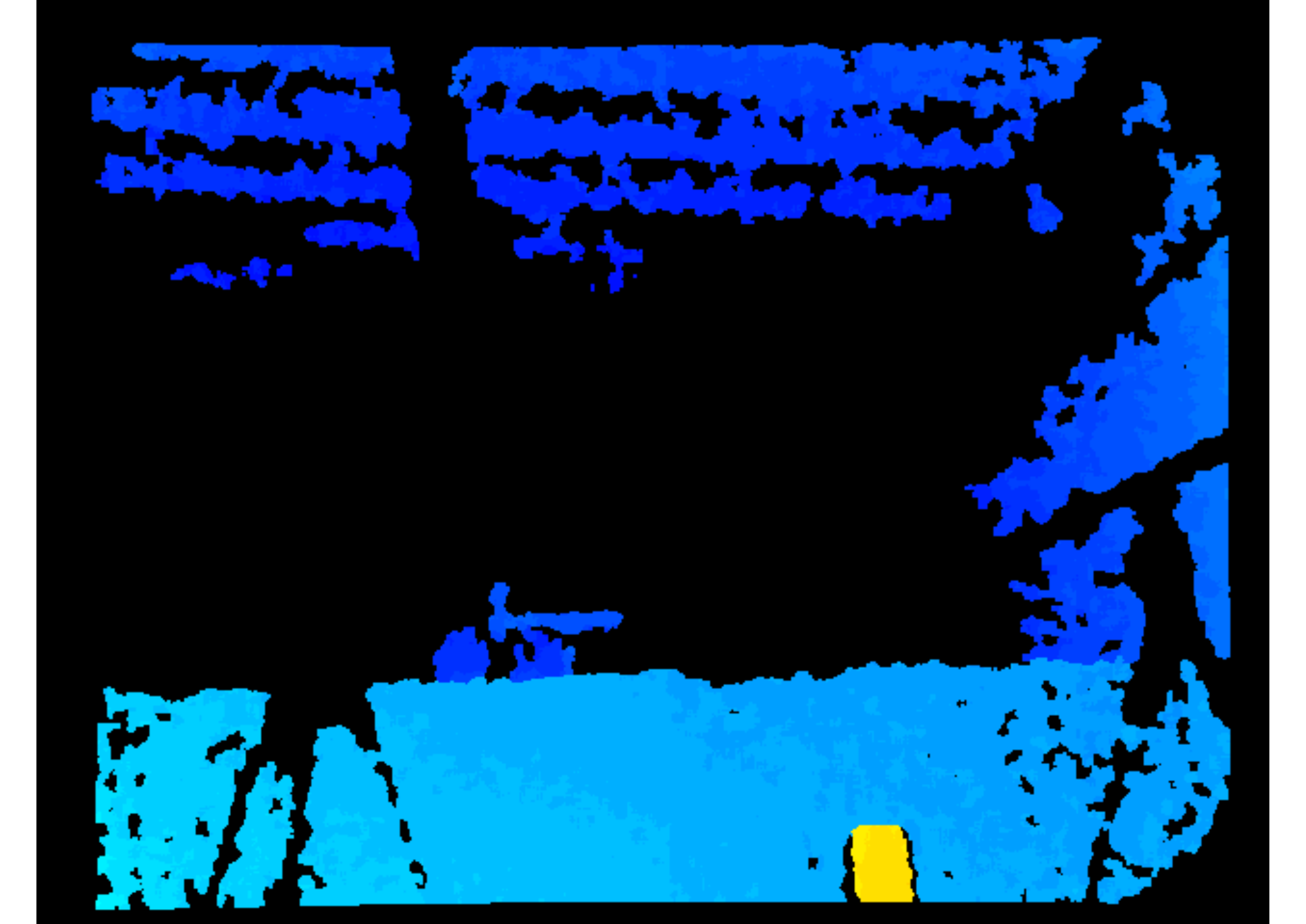} &
		\includegraphics[scale=0.183]{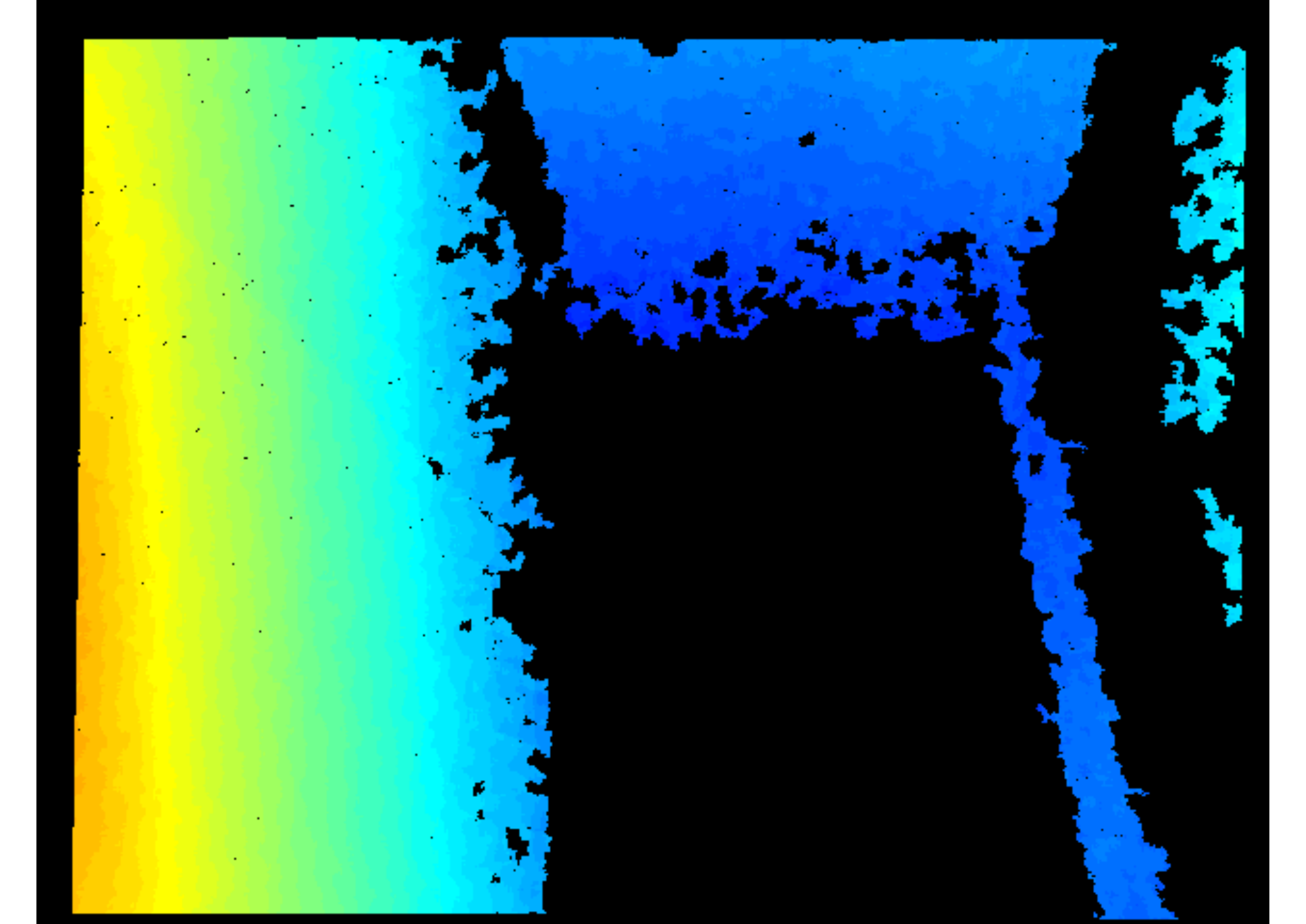} &
		\includegraphics[scale=0.183]{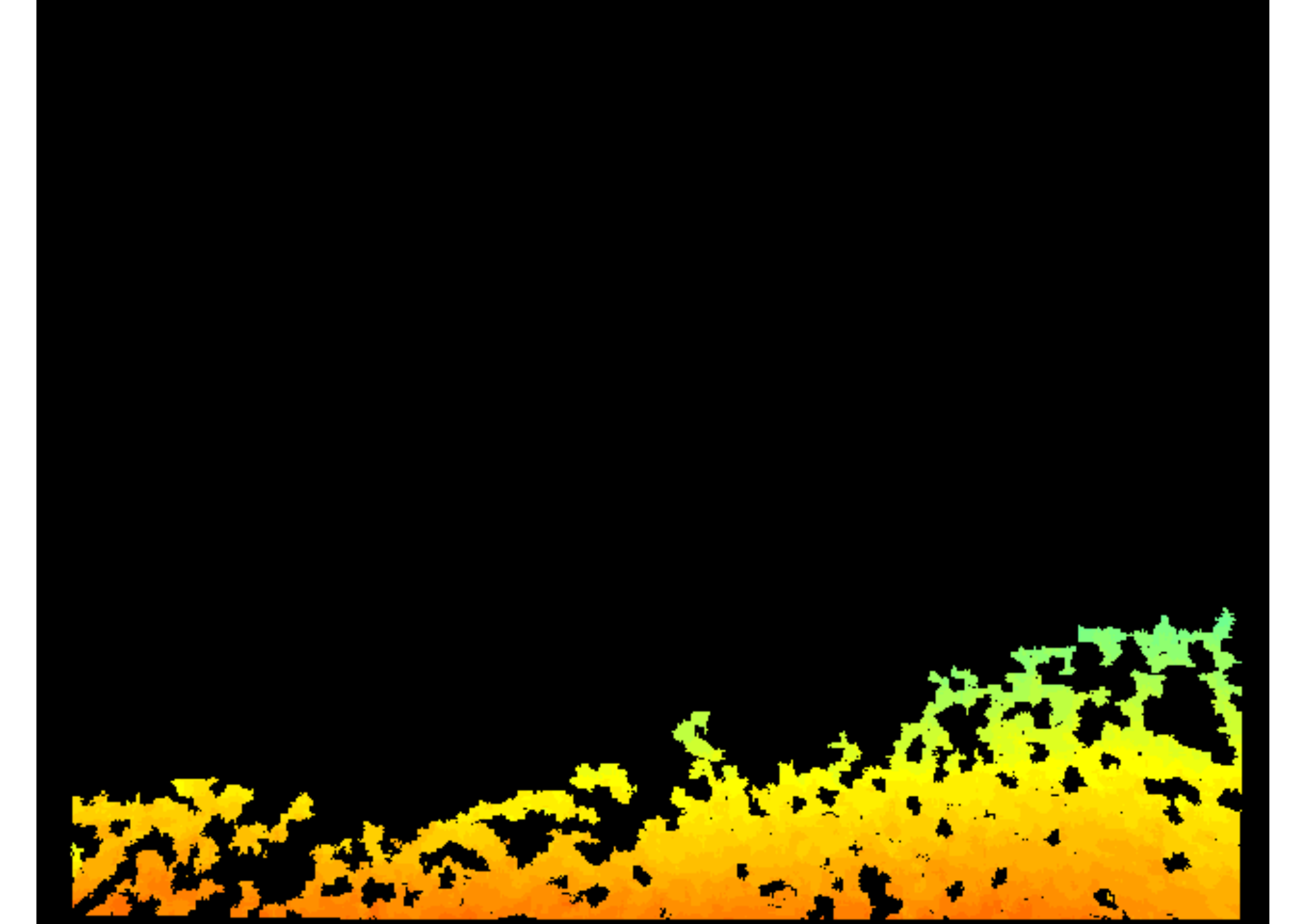} \\
		(a) fr2/360\_hemisphere & (b) fr2/360\_hemisphere & (c) foyer & (d) park
	\end{tabular} 
	\caption{Examples of the sequences used in our experiments. Top: Inverse depth of features obtained by either the depth map or the depth estimation process. Middle: Inverse depth maps obtained by the depth map registration process. Bottom: Original inverse depth maps captured by the depth camera.}
	\label{fig8}
	\vspace*{-2mm} 
\end{figure*}

In this dataset, mode B yields generally lower errors than mode A. This can be explained by the existence of spurious depth measurements due to NIR interference (shown in Fig. \hyperref[fig8]{\ref*{fig8}.d}) and non-Lambertian effects (shown in Fig. \ref{fig7}). These depth errors are further propagated and accumulated by the depth
map registration, which can affect severely the pose estimation, as shown in Fig. \ref{fig7}, especially for Mode A, which relies more on the depth measurements.

{\renewcommand{\arraystretch}{1.2}
\begin{table}[]
\centering
\caption{Performance of SPLODE on author-collected sequences reported as the final error of the trajectory and the number of frames where pose estimation is declared to fail.}
\label{tab:surrey_results}
\begin{tabular}{|c|c|c|c|c|c|}
\hline
\multicolumn{1}{|l|}{\multirow{2}{*}{\begin{tabular}[c]{@{}l@{}}Sequence\\ (Length)\end{tabular}}} & \multirow{2}{*}{\begin{tabular}[c]{@{}c@{}}Depth \\ estimates\end{tabular}} & \multirow{2}{*}{Mode} & \multicolumn{3}{c|}{SPLODE} \\ \cline{4-6} 
\multicolumn{1}{|l|}{} &  &  & Points & Lines & All \\ \hline
\multirow{3}{*}{\begin{tabular}[c]{@{}c@{}}Foyer\\ (65 m)\end{tabular}} & No & - & \begin{tabular}[c]{@{}c@{}} Fail\end{tabular} & \begin{tabular}[c]{@{}c@{}}Fail\end{tabular} & \begin{tabular}[c]{@{}c@{}}4.4 m\\ 3 losses\end{tabular} \\ \cline{2-6} 
 & \multirow{2}{*}{Yes} & A & \begin{tabular}[c]{@{}c@{}}2.3 m\\ 11 losses\end{tabular} & \begin{tabular}[c]{@{}c@{}}2.8 m \\ 0 losses\end{tabular} & \begin{tabular}[c]{@{}c@{}}1.2 m \\ 0 losses\end{tabular} \\ \cline{3-6} 
 &  & B & \begin{tabular}[c]{@{}c@{}}2.6 m\\ 19 losses\end{tabular} & \begin{tabular}[c]{@{}c@{}}1.7 m\\ 0 losses\end{tabular} & \begin{tabular}[c]{@{}c@{}}0.9 m\\ 0 losses\end{tabular} \\ \hline
\multirow{3}{*}{\begin{tabular}[c]{@{}c@{}}Park\\ (76 m)\end{tabular}} & No & - & \begin{tabular}[c]{@{}c@{}}4.9 m\\ 38 losses\end{tabular} & Fail & \begin{tabular}[c]{@{}c@{}}7.0 m \\ 5 losses\end{tabular} \\ \cline{2-6} 
 & \multirow{2}{*}{Yes} & A & \begin{tabular}[c]{@{}c@{}}3.5 m \\ 0 losses\end{tabular} & \begin{tabular}[c]{@{}c@{}}Fail\end{tabular} & \begin{tabular}[c]{@{}c@{}}2.8 m \\ 0 looses\end{tabular} \\ \cline{3-6} 
 &  & B & \begin{tabular}[c]{@{}c@{}}2.0 m\\ 0 losses\end{tabular} & \begin{tabular}[c]{@{}c@{}}Fail\end{tabular} & \begin{tabular}[c]{@{}c@{}}1.8 m \\ 0 losses\end{tabular} \\ \hline
\end{tabular}
\end{table}
}

\section{Conclusions}
This paper presents an RGB-D based odometry method that achieves robustness to poorly textured scenes and depth sensor failures by combining point and line features, and depth sensor measurements with temporal stereo. \par
Our results show that the depth estimation framework introduced by this method is beneficial in large indoor environments (e.g. warehouses) and outdoor environments, where the depth information captured by depth cameras is too sparse. Moreover, our experiments indicate that gross depth errors occur typically outdoors, due to NIR interference with the sunlight, and its impact on the pose estimation can be minimized by adopting a more active depth estimation approach (referred to as Mode B) than just estimating depth when measurements are missing (Mode A). However, because Mode B is more computational intensive than Mode A, adaptive switching between both is a worthwhile research direction. Another possible improvement of the system, is to make the pose estimation also probabilistic by weighting the residuals according to the uncertainties of both depth measurements and estimates: In this work, SPLODE relies on a finely-tuned maximum uncertainty threshold to accept sufficiently precise depth estimates, but by using the depth estimate uncertainty in the pose estimation, such threshold could be further relaxed to exploit more features with higher uncertainties.


\bibliographystyle{ieeetr} 
{\footnotesize
\bibliography{iros_ref}
}

\end{document}